\documentclass[11pt,a4paper]{article}
\usepackage{times,latexsym}
\usepackage{url}
\usepackage[T1]{fontenc}
\usepackage{graphicx}
\usepackage{array}
\usepackage{caption} 
\usepackage{multirow} 
\usepackage{paralist, tabularx}
\usepackage[acceptedWithA]{tacl2021v1} 
\usepackage{diagbox}
\newcommand\blfootnote[1]{%
  \begingroup
  \renewcommand\thefootnote{}\footnote{#1}%
  \addtocounter{footnote}{-1}%
  \endgroup
}
\usepackage{xspace,mfirstuc,tabulary}

\newif\iftaclinstructions
\taclinstructionsfalse 
\iftaclinstructions
\renewcommand{\confidential}{}
\renewcommand{\anonsubtext}{(No author info supplied here, for consistency with
TACL-submission anonymization requirements)}
\newcommand{\instr}
\fi

\iftaclpubformat 

\else

\fi

\newcommand{\cptname}{WangchanLION-V3-8B }

\title{Mangosteen: An Open Thai Corpus for Language Model Pretraining}

\author{
Wannaphong Phatthiyaphaibun\textsuperscript{$\spadesuit,\dagger$},
Can Udomcharoenchaikit\textsuperscript{$\spadesuit,\dagger$},
Pakpoom Singkorapoom\textsuperscript{$\spadesuit$}, \\
\textbf{Kunat Pipatanakul}\textsuperscript{$\clubsuit$}, 
\textbf{Ekapol Chuangsuwanich}\textsuperscript{$\diamondsuit$}, \\
\textbf{Peerat Limkonchotiwat}\textsuperscript{$\heartsuit,\ddagger$},
\textbf{Sarana Nutanong}\textsuperscript{$\spadesuit,\ddagger$} \\
  \textsuperscript{$\spadesuit$}Vidyasirimedhi Institute of Science and Technology 
  \textsuperscript{$\clubsuit$}SCB10X \\
  \textsuperscript{$\diamondsuit$}Chulalongkorn University 
  \textsuperscript{$\heartsuit$}AI Singapore
}

\date{}

\begin{document}
\maketitle
\blfootnote{\textsuperscript{$\dagger$}Co-first Authors, \textsuperscript{$\ddagger$}Corresponding Authors}
\begin{abstract}

Pre‑training data shapes a language model’s quality, but raw web text is noisy and demands careful cleaning. 
Existing large‑scale corpora rely on English‑centric or language‑agnostic pipelines whose heuristics do not capture Thai script or cultural nuances, leaving risky material such as gambling content untreated. 
Prior Thai-specific efforts customize pipelines or build new ones, yet seldom release their data or document design choices, hindering reproducibility and raising the question of how to construct a transparent, high‑quality Thai corpus.
We introduce Mangosteen: a 47 billion‑token Thai corpus built through a Thai-adapted Dolma pipeline that includes custom rule‑based language ID, revised C4/Gopher quality filters, and Thai‐trained content filters, plus curated non‑web sources such as Wikipedia, Royal Gazette texts, OCR‑extracted books, and CC‑licensed YouTube subtitles.
Systematic ablations using GPT-2 show the pipeline trims CommonCrawl from 202M to 25M documents while raising SEA-HELM NLG from 3 to 11; an 8B-parameter SEA-LION model continually pre‑trained on Mangosteen then surpasses SEA-LION-v3 and Llama-3.1 by about four points on Thai benchmarks.
We release the full pipeline code, cleaning manifests, corpus snapshot, and all checkpoints, providing a fully reproducible foundation for future Thai and regional LLM research.
\footnote{\href{https://huggingface.co/collections/aisingapore/wangchanlion-v3-687a362d8f0ea2fe4077c6b3}{All artifacts in this papers}}



\end{abstract}

\section{Introduction}
Pre-training datasets are an important part of training a language model. The size, the diversity, and the quality of the pre-training data play a crucial role in obtaining a high-performing language model.
These datasets are compiled by scraping vast amounts of web text, augmented with diverse sources like books, scholarly articles, and code.
%
%
These datasets are enormous, often reaching hundreds of billions of tokens.
Raw web corpora are messy, containing noise, duplicates, and harmful content. If not carefully filtered, these issues can lead to undesirable model behavior.
To improve the quality of the dataset, a filtering pipeline is needed to remove these unwanted samples.

Most large-scale English pre-training datasets, such as CC-100~\cite{wenzek-etal-2020-ccnet}, C4~\cite{10.5555/3455716.3455856}, RefinedWeb~\cite{penedo2023refinedwebdatasetfalconllm}, and Dolma~\cite{soldaini-etal-2024-dolma}, are mainly derived from internet data such as the Common Crawl corpus\footnote{\url{https://commoncrawl.org/}}. 
Hence, this requires extensive cleaning and filtering.
%
%
%
However, most data-cleaning pipelines for web pages are optimized for the English language using known best practices.
%
For multilingual datasets that support Thai, such as CC-100, FineWeb2~\cite{penedo2024fineweb-2}, and HPLT v2~\cite{burchell2025expandedmassivemultilingualdataset}, the data cleaning pipelines are usually language-agnostic. 
Although being language-agnostic can lead to a generalized design for a data cleaning pipeline, this approach prioritizes broad applicability over incorporating local, cultural, and language-specific knowledge. 
Consequently, language-specific filtering steps are necessary to prevent unwanted information from being used to train language models.
For example, the FineWeb2 dataset contains content from gambling websites, which are illegal in Thailand. 

For language-specific datasets, many data collection projects utilize existing data cleaning pipelines by customizing them to make these pipelines more suitable for their target languages.
This ranges from making a small modification in one process, for instance, the Latxa project~\cite{etxaniz-etal-2024-latxa} for Basque, which adapted the Dolma and Corpus Cleaner v2 pipelines with a specific change in the filtering step, to customizing the whole pipeline. 
Examples of more extensive customization include the FinGPT~\cite{luukkonen-etal-2023-fingpt} and IndicLLMSuite~\cite{khan-etal-2024-indicllmsuite} projects that built their own dedicated pipelines for Finnish and Indian languages, respectively.
For the Thai language, the approach ranges from using a standard language-agnostic pipeline, such as the Typhoon-2 project, which uses Trafilatura~\cite{barbaresi-2021-trafilatura} and text-dedup~\cite{chenghao_mou_2023_8364980} directly while creating its own heuristic filtering, to customizing the whole pipeline to specifically handle Thai. 
An example of a dedicated Thai data cleaning pipeline is from the OpenThaiGPT~\cite{yuenyong2025openthaigpt15thaicentricopen}, in which they build the data preprocessing pipeline from scratch.
However, these projects focus on open models rather than the openness of their training data and collection methods. 
While this provides a valuable resource, it means the pre-training corpora remain inaccessible and the pipeline's design choices and empirical backing are not detailed, which is an important consideration for researchers focused on reproducibility.
This raises a central question: \emph{What does it take to build a high-quality pre-training corpus for a language like Thai, and how should existing pipelines be adapted to meet its linguistic and cultural specifics?}

To address the lack of Thai pre-training resources and to increase the exposure of Thai within the open-source NLP community, we have developed Mangosteen, a large-scale pre-training dataset for the Thai language. This dataset contains 47.4 billion tokens and is available under an open-source license. In addition, we provide an in-depth analysis of our data cleaning pipeline, designed specifically for Thai, including an ablation study that confirms the effectiveness of each processing step. 
We customized the Dolma~\cite{soldaini-etal-2024-dolma} pipeline to curate variants of Thai Common Crawl data through a systematic cleaning process because the Dolma pipeline is easy to customize and supports parallel processing. For language identification, we developed a rule-based script to accurately detect Thai content. The quality filter was adapted by modifying the C4 and Gopher filter rules to address specific Thai language nuances. Furthermore, we implemented a content filter composed of two components: an obscene content filter and a gambling-related content filter, both of which were trained on Thai-specific data to effectively remove undesirable content. 
In addition to Common Crawl data, we include data from several other sources to ensure diversity. These include Wikipedia, YouTube subtitles, and even text extracted from open-access books using OCR.


To determine the most effective data pipeline for the Thai language, we systematically evaluate the impact of each processing step on language model performance.
We test each step in our pipeline to find the optimal settings for the Thai language. This process involves pre-training a GPT-2 model~\cite{radford2019language} with 124M parameters on 10B tokens for each configuration. The effectiveness of each step is then confirmed by evaluating the trained model against benchmarks like SEACrowd and SEA-HELM.
Compared to the uncleaned Common Crawl data, data processed through our pipeline achieves better performance despite being multiple times smaller. Furthermore, we can improve the already-cleaned FineWeb2 data by passing it through our pipeline, resulting in a much smaller dataset that maintains a similar level of performance.
Furthermore, to confirm the robustness of our data, we developed a new Thai large language model by further pre-training the Southeast Asian model (SEA-LION~\cite{aisingapore-sealion-3-1}) on our dataset.
Our LLM outperforms other Llama3.1-based models on both the Thai LLM Leaderboard and SEA-HELM benchmarks.
%

%

We summarize the contribution of our paper as follows:
\begin{compactitem}[\hspace{3mm}•]
    %
    \item We introduce a data cleaning pipeline for improving the quality of data, which adapts the Dolma pipeline for the language's specific needs. 
    We also describe the customization steps taken in this process. 
    Moreover, we conduct an ablation study to demonstrate the effectiveness of our data cleaning pipeline.

    \item 
    Beyond Common Crawl data, through an extensive curation effort, we curate and extract high-quality texts from numerous other sources, including Wikipedia, YouTube subtitles, open-access books whose text we extracted via OCR, official documents from the Royal Gazette, open government data, established resources like Wikipedia, and existing Thai datasets on Hugging Face.
    
    \item In a large-scale effort to advance open data and open models for the Thai language, we introduce a high-quality named ``\textbf{Mangosteen}'', 47B token pre-training corpus. To demonstrate its effectiveness, we also present WangchanLION-V3-8B, a new, fully open-source Thai LLM, developed by further pre-training a SEA-LION-based model on our dataset.  In line with our commitment to open science, this entire ecosystem—the dataset, the model, and all related code—is available under permissive open-source licenses.
\end{compactitem}

\section{Related Work}






\subsection{Text Pretrained Corpora}




A critical design choice for any pre-training corpora is data cleaning, which ensures the quality of the data and its effectiveness for the pre-training process.
Prior efforts~\cite{wenzek-etal-2020-ccnet,10.5555/3455716.3455856, xue-etal-2021-mt5,pile} focusing on improving data quality using simple techniques such as language identification, document deduplication, and quality filtering using a perplexity filter model on web corpora, where the language is monolingual (English only) to multilingual (more than 100 languages).
Later works also demonstrate more pristine text processing using another model to classify text quality~\cite{ortiz-suarez-etal-2020-monolingual}. 
The C4 corpus~\cite{raffel2023exploringlimitstransferlearning} was created by the rule, which was later named the ``C4'' rule.  
Gopher rule~\cite{Rae2021ScalingLM}, and more complicated data rules, e.g., RefinedWeb~\cite{penedo2023refinedwebdatasetfalconllm} demonstrate the pipeline that uses URL filtering, text extraction, language identification, repetition removal, document-wise filtering, Line-wise corrections, fuzzy deduplication, and exact deduplication, outperform previous data cleaning processes. 
%
%

Recently, the size of the pre-training corpora has been increased significantly according to the model size, which has been increased from 100 million parameters (i.e., BERT-base~\cite{DBLP:journals/corr/abs-1810-04805}) to a large language model (i.e., Llama-8B~\cite{grattafiori2024llama3herdmodels}).
Therefore, most recent works propose a large-scale pre-training dataset with methods for creating clean and high-quality pre-training data to improve the pre-training model's performance.
The Dolma corpus~\cite{soldaini-etal-2024-dolma} has been an English-pretrained dataset of 3 trillion tokens mixed with many sources: web pages, academic publications, code, books, and encyclopedic materials. 
They released a dataset and a pipeline to create dataset transparency in the language model, where the pipeline includes language filtering, URL and text overlap deduplications, quality filters, and content filters.
FineWeb~\cite{penedo2024finewebdatasetsdecantingweb} is clean and deduplicates English web data from CommonCrawl, where this previous work uses a linear regression model to classify educational texts that are made using Llama-3 synthetic annotations. 
FineWeb2~\cite{penedo2024fineweb-2} creates over 1000 languages for pre-training models. Its methodology is similar to FineWeb, but data is deduplicated per language globally, and C4 filters are not used. 
%

In general, these pipelines are similar and mostly open-source, but the majority do not include instructions on how to apply them to other languages.
Furthermore, the aforementioned pipelines fail to explain the processes that make the preprocessing steps effective for the target language, and these pipelines are not effectively modified to an extent that makes them suitable for the target language. 
For instance, some languages necessitate either language-specific word segmentation tools or a newly trained model to filter out offensive content, including pornography and gambling material.

\subsection{Thai-text pre-training datasets and models} \label{subsec:thai_pre-training}

The most prominent pre-training datasets for Thai are typically included as subsets within multilingual corpora. 
Early examples include the multilingual BERT~\cite{DBLP:journals/corr/abs-1810-04805} model, which uses Wikipedia and includes Thai content. 
This is followed by the CC-100 corpus~\cite{wenzek-etal-2020-ccnet}, which supports Thai and is used in models such as XLM and XLM-R~\cite{conneau-etal-2020-unsupervised}. 
The OSCAR corpus~\cite{ortiz-suarez-etal-2020-monolingual} also includes Thai and forms part of the dataset used in GPT-2 Thai by Flax's GPT-2 base~\footnote{\url{https://huggingface.co/flax-community/gpt2-base-thai}}. 
More recently, trillion-scale multilingual pre-training datasets such as CulturaX~\cite{nguyen2023culturaxcleanedenormousmultilingual}, FineWeb2~\cite{penedo2024fineweb-2}, and HPLT v2~\cite{burchell2025expandedmassivemultilingualdataset} have also included Thai among hundreds of supported languages.
%
%
%
%
However, their language-agnostic approach disregards language-specific and local knowledge vital for accurate data cleansing.
Moreover, the Thai community also proposes a Thai CPT project to improve the number of data and the model's robustness.
OpenThaiGPT~\cite{yuenyong2025openthaigpt15thaicentricopen} is a project that performs continual pre-training on Llama models with Thai language by extending vocabulary and continually pre-training Thai language datasets.
SambaLingo~\cite{csaki2024sambalingoteachinglargelanguage} is a Llama-2 model that does vocabulary expansion and continuous pre-training for nine languages, including Thai, using data sourced from CulturaX~\cite{nguyen2023culturaxcleanedenormousmultilingual}.
Typhoon~\cite{pipatanakul2023typhoonthailargelanguage} performs CPT on cleaned and deduplicated CC and extends vocabulary with Mistral-7B, and Typhoon 2~\cite{pipatanakul2024typhoon2familyopen} applies CPT on Qwen 2.5 7B and Llama 3.1 by combining the original pipeline with a multi-classifier model for data filtering.
OpenThaiLLM-Prebuilt-7B\footnote{\url{https://medium.com/nectec/openthaillm-prebuilt-release-f1b0e22be6a5}} is a project that continues pre-training on Qwen 2 with the Thai and Chinese datasets.

Although previous works have demonstrated the possibility of improving Thai's performance using CPT, they only partially open-sourced their pipelines or did not study the effect of each individual component.
For example, Typhoons use heuristic filtering like RefineWeb~\cite{penedo2023refinedwebdatasetfalconllm} and DCLM~\cite{li2025datacomplmsearchgenerationtraining}, but do not open-source their pipeline code or model to enable reproducibility.
OpenThaiGPT open-sourced their data processing pipeline\footnote{\url{https://github.com/OpenThaiGPT/data-processing}}; however, they do not provide thorough study of the pipeline, including the necessary analyses and resources required for full reproducibility. For example, they do not share the dataset or the model used for perplexity-based filtering.
Therefore, the Thai research community is struggling to develop or make any forward progress without open-source data and a study on its design decision.

\subsection{Gap Summary}

Existing data cleaning pipelines like C4, RefinedWeb, and Dolma, while scalable, are designed primarily for English or applied in a language-agnostic way, requiring further adjustments to fit distinct scripts and cultural norms. 
Thai pre-training efforts, such as OpenThaiGPT and Typhoon, reuse such pipelines or apply ad hoc modifications without empirical validation, and rarely release their datasets or pipelines, limiting reproducibility. 
This paper addresses these gaps by systematically applying the principles found in C4, RefinedWeb, and Dolma to the Thai-specific context. 

\section{Data Curation}

\label{sec:data_curation}
The goal of data curation is to build a Thai pre-training corpus that covers the broadest possible set of domains, enabling adaptations in various downstream tasks.
%
%
We collect datasets from various sources that can be classified into two categories: (i) Common Crawl-Derived Dataset and (ii) Curated Non-Common Crawl. We have a total of 30.1M documents and 47.4B tokens, as shown in Table~\ref{tab:document_stats}.

\begin{table}[h!]
    \centering
    \scalebox{0.8}{
    \begin{tabular}{|l|l|l|}
        \hline
        \textbf{Source} & \textbf{Documents}& \textbf{Tokens (B)} \\
        \hline
        Common Crawl-Derived & 29.7M& 45.9\\
        Curated Non-Common Crawl & 425,304 & 1.5 \\
        \hline
        Total & 30.1M & 47.4 \\
        \hline
    \end{tabular}}
    \caption{Document and token statistics (using the Llama 3 tokenizer)}
    \label{tab:document_stats}
\end{table}

\subsection{Common Crawl-derived Dataset}

Following the best practices from other pre-training corpora~\cite{raffel2023exploringlimitstransferlearning,Rae2021ScalingLM,penedo2023refinedwebdatasetfalconllm,soldaini-etal-2024-dolma}, our Common Crawl-derived data consists of two main sources: raw Common Crawl snapshots and the FineWeb2 dataset. We focus specifically on the Thai language subset from both sources.

\noindent
\textbf{Common Crawl}
%
We collect Common Crawl dataset by processing each Common Crawl snapshot from 2018-30 to 2023-23 and extract only Thai-language content using Common Crawl metadata.
For text extraction, we use trafilatura ~\citep{barbaresi-2021-trafilatura}. Followed by our data cleaning pipeline (see Section \ref{sec:data_processing} for full details) to filter and deduplicate the result.


\noindent
\textbf{FineWeb2}
To further increase the quantity of our Common Crawl-derived dataset, we incorporate FineWeb2, a large-scale pre-training corpus built from Common Crawl data, which employs multiple filtering techniques to improve dataset quality. 
FineWeb2 spans from summer 2013 to April 2024. The Thai subset contains approximately 51.4 billion words across 35 million documents. We further process the cleaned FineWeb2 dataset using our data cleaning pipeline (Section \ref{sec:data_processing}) to deduplicate overlapping URLs and text inherited from the Common Crawl source while also enhancing language-specific quality. After this step, the final cleaned Thai FineWeb2 subset consists of around 7.3 billion words and 4.6 million documents.


\subsection{Curated Non-Common Crawl Dataset}

In addition to the Common Crawl-based data, we enhance the data diversity through the efforts of incorporating harder-to-reach data as follows. 
%
%
%
%
(1) A significant subset of our source documents consisted of scanned or image-based PDFs, whose textual content is not directly accessible. To overcome this, we perform Optical Character Recognition (OCR) to extract the text from these files using Marker\footnote{\url{https://github.com/VikParuchuri/marker}}. Due to the diverse publication formats of the scanned documents, we must customize the OCR pipeline for each document format and clean them manually.
(2) We also extract more text from YouTube video subtitles using the provided metadata, so we can create a pipeline that filters suitable Thai video subtitles that use the Creative Commons (CC) license. To extract subtitles, we modified Scrapetube's source code\footnote{\url{https://github.com/dermasmid/scrapetube}} to exclusively retrieve Creative Commons licensed videos. Our two-step methodology involved:  
1. Compiling Thai keywords across domains (mathematics, investing, fitness, gaming, films, etc.), applying CC/subtitle filters to gather $\leq$1,000 video metadata entries per keyword.  
2. Using quoted channel names as queries to identify predominantly CC-licensed content. 
Video IDs were processed via YouTube's Transcription API\footnote{\url{https://github.com/jdepoix/youtube-transcript-api}} to generate textual JSON Lines. As a final validation step, non-CC contents were removed via yt\_dlp\footnote{\url{https://github.com/yt-dlp/yt-dlp}}.
%

%
The curated non-Common Crawl dataset spans over a large number of sources, which covers the following categories:

\begin{compactitem}[\hspace{3mm}•]
    \item \textbf{Encyclopedic}: We collect Thai data from Wikipedia, Wikibooks, Wikiquote, and Wikisource, then use WikiExtractor~\cite{Wikiextractor2015} 
    to extract text and then remove the HTML tags and adjust the formatting according to the WikiExtractor tool's outputs. 
    \item \textbf{Finance}: We use the Financial Text Data Collection\footnote{\url{https://huggingface.co/datasets/airesearch/CMDF_VISTEC}} by the VISTEC-depa Thailand artificial intelligence research institute. This dataset includes financial reports from Thai companies and Thai financial news.
    \item \textbf{Legal}: We collect Thai legal data from multiple HuggingFace repositories, including acts and statutes, constitutions, and Creative Commons licenses.
    \item \textbf{Government}: We collect data that is publicly released by government agencies from multiple sources. This includes the Open Government Data of Thailand, which is a Thai government effort to improve transparency and engagement through access to open data.  Government data often comes in scanned PDF format, which requires extraction through OCR. Furthermore, we have to extract texts from various formats such as DOCX and CSV.  
    \item \textbf{Education}: We collect educational materials ranging from informative articles and classical Thai literature to advanced academic research.
    \item \textbf{YouTube}: We collect Thai YouTube subtitles from videos with the CC BY 3.0 license. 
\end{compactitem}

%


%
%

%



%
 Table~\ref{tab:noncc-domain-count} shows the document count distribution for each category.
The curated Non-Common Crawl corpus comprises 425,304 Thai-language documents from sources not present in Common Crawl. 
After deduplication, we split the corpus into a training set of 397,488 documents and a validation set of 4,044 documents, where the Non-Common Crawl dataset was curated from a wide variety of web sources and includes data from encyclopedic websites, financial reports, education, legal corpora, academic literature, and YouTube transcripts. 
In total, the Non-Common Crawl corpus contributes an additional 1.54 billion tokens to the dataset, tokenized using the \texttt{meta-llama/Llama-3.2-1B} tokenizer.
All sources were verified to ensure the use of permissive content licenses. 
The full list of text sources can be found in \url{https://github.com/vistec-AI/Mangosteen}.
%

\begin{table}[h!]
\centering
\scalebox{0.9}{
\begin{tabular}{| l | r | r |}
\hline
Domain & {Count} & {Percentage} \\
\hline
Encyclopedic & 166,187 & 41.34\\
\hline
Finance & 86,813 & 21.59\\
\hline
Government & 72,879 & 18.13\\
\hline
Legal & 52,343 & 13.02\\
\hline
YouTube & 17,837 & 4.43\\
\hline
Education & 5,911 & 1.47\\
\hline
\end{tabular}}
\caption{Document counts and percentage in Non-Common Crawl by domain.}
\label{tab:noncc-domain-count}
\end{table}
A more detailed table on sources of information for the Non-Common Crawl dataset can be seen in the appendix.
We also use a deduplication process to ensure this new data does not overlap with the web data collected earlier.
By incorporating this non-Common Crawl into our data, we gain an additional 1.5B tokens of high-quality data.
Although these data are of high quality, we still need to clean the noisy web data (e.g., Common Crawl and Fineweb2) to ensure the overall quality of our collected dataset.
%
%
Therefore, we still require a data cleaning pipeline to mitigate these problems.



\section{Data cleaning pipeline for web data}
\label{sec:data_processing}
\subsection{Overview}

%
A data cleaning pipeline is a crucial component for ensuring the quality of the pre-training data.
For the Thai NLP community, previous Thai LLMs also demonstrate how to gather a large-scale dataset for pre-training an LLM.
%
However, as we discuss in Section~\ref{subsec:thai_pre-training}, the focus of existing works has been on the model release rather than the training data and collection methods. 
Consequently, the pre-training corpora often remain inaccessible, and pipeline design choices are not discussed in detail, which is an important consideration for researchers focused on reproducibility and adaptability.

In this work, we propose a novel data collection and cleaning pipeline tailored to the unique characteristics and challenges of Thai data. 
%
In particular, we present a Thai adaptation of the well-known Dolma~\cite{soldaini-etal-2024-dolma} data curation pipeline, featuring a new data processing design customized for the Thai language.
After applying our data filtering pipeline, we obtained a large-scale dataset containing 45.9B tokens.
The Dolma pipeline consists of four steps: language identification, quality filters, deduplication, and content filters, described as follows.

    
    

    


\subsection{Language Identification}
The first step of our data preprocessing is language identification, where we remove all non-Thai data from the pre-training corpus. 
In particular, we discard any document containing text in other languages, such as Lao or Burmese.

Dolma used FastText~\cite{joulin-etal-2017-bag} as a language identifier for English corpora.
However, we found that when we compare three language identifiers: langdetect, lingua, and the FastText language identifier model, all three failed to perform this task on Thai text, as shown in the Appendix~\ref{appendix:lang_ident}.
Therefore, we use a simple and efficient rule-based approach that identifies Thai text with a regular expression based on the Thai Unicode character range.
This filter allows only documents in which at least half of the characters are Thai Unicode characters to pass.
%
%

\subsection{Quality Filters}
\label{sec:quailty_filters}

As a core step in our pipeline, we filter out low-quality data. To do this, we follow the approach used by Dolma, applying quality filtering rules such as C4~\cite{raffel2023exploringlimitstransferlearning} and Gopher~\cite{Rae2021ScalingLM}.
Moreover, we also customize these rules to better account for the unique characteristics of the Thai language and to integrate insights from our observations.

\noindent
\textbf{C4}:
We adopt all C4 rules from Dolma as follows:
\begin{compactitem}
\item Contains curly braces (e.g., \{ or \})
\item Includes the placeholder text “lorem ipsum”
\item Contains JavaScript code or references
\item Includes offensive or inappropriate language (using a filter we built specifically for Thai).
\item Has lines that lack ending punctuation
\item Contains lines with fewer than three words
\end{compactitem}

To detect offensive content, we employ a custom Thai lexicon. Because Thai text is not delimited by spaces, we segment it using the nlpO3 tokenizer~\cite{suntorntip_2024_14082449} from PyThaiNLP~\cite{phatthiyaphaibun-etal-2023-pythainlp}. We also implement a rule to remove Unicode replacement characters, as these are unknown or unrepresentable characters.
Although we retain all original Dolma rules, we omit the c4\_no\_punc rule, since Thai sentences do not conventionally end with punctuation such as a period.

\noindent
\textbf{Gopher}:
The Dolma pipeline uses Gopher rules as part of its quality filtering process for English text. However, based on our analysis of the Common Crawl-based data, we modified some of these rules and adjusted their corresponding thresholds to make them suitable for Thai.
\begin{compactitem}
\item First, we raise the minimum document length from 50 to 200 words and cap it at 100,000 words, since Thai sub-200-word texts in our Common Crawl corpus of 200,000 examples were predominantly low quality or sourced from pornographic and gambling sites.
\item Second, we exclude any document in which Thai consonants account for less than 80\% of all characters, reflecting our bespoke characteristic of the Thai language. 
\item Next, we discard texts where over 30 \% of lines end with an ellipsis (``…'' or three dots). Ellipses are often used to mark the end of incomplete snippets of an article or excerpt. 
\item We also introduce a new rule to filter out any document containing markers of truncated content in Thai, e.g., “continue reading”, “read more” or “read more at”, and change the list of stop words to Thai.
\end{compactitem}

To justify the higher minimum document length threshold, we compared WangchanBART~\footnote{\url{airesearch/wangchanbart-base}} perplexity statistics at lower bounds of 50 versus 171 words and found both mean and median perplexities to be lower at the higher threshold; therefore, we set the bound at 200 words. Finally, to speed up processing, we employ the ICU tokenizer rather than the slower nlpO3 tokenizer.

\subsection{Deduplication}

\paragraph{Deduplication by URL.}
%
A common text preprocessing step is the removal of duplicate content. For this, we utilize the deduplication technique from the Dolma pipeline~\cite{soldaini-etal-2024-dolma}, which uses a Bloom filter to identify duplicate URLs. 
We found this method to be effective on our Thai web dataset and therefore kept this step of the pipeline unmodified.

\paragraph{Deduplication on text overlap.}
%
%
We perform document-level deduplication using the Dolma Bloom filter. Dolma's paragraph-level deduplication is ineffective for Thai web data since the \texttt{UTF-8} newline (\texttt{\textbackslash n}) is not always a good indicator of the paragraph boundary in Thai text.
Therefore, we only apply deduplication at the document level.

\subsection{Content Filters}

\label{sec:content_filters}
%
Following Dolma's approach, we use FastText for language-specific filtering due to its excellent speed-accuracy balance for large-scale data cleaning. 
We employ pre-trained FastText vectors for 157 languages~\cite{grave-etal-2018-learning} to train filters for adult and gambling content.
%
%
To train our two content filters, we create dedicated datasets for binary classifiers. We label documents as belonging to a specific class if they contain three or more distinct words from a predefined list for that class. For gambling content, we also include documents that an LLM identified as promoting gambling. Finally, we randomly sample a subset of the data for human validation, ensuring the quality of our content filtering datasets.
Moreover, we changed the phone number rule for the Personally Identifiable Information (PII) filter to ensure Thai phone number compatibility.

\section{Experimental Settings}

\subsection{CPT and SFT Details}
\label{sec:cpt_setting}
To assess the effectiveness of our pre-training data (47.4B tokens as shown in Section~\ref{sec:data_curation}), we use it for continuous pre-training (CPT) on SEA-LIONv3-Llama-8B-instruction~\cite{aisingapore-sealion-3-1}. 
We use the training setup as follows:
\begin{compactitem}
    \item max\_seq\_len: 8192
    \item learning rate: 5.0e-6 
    \item optimizer: decoupled\_lionw
    \item lr\_scheduler\_type: cosine\_with\_warmup
    \item num\_train\_epochs: 1
    \item GPU: H100 (64 GPUs)
    \item Time: 1d 12h 24m
\end{compactitem}
After CPT, we conduct supervised fine-tuning (SFT) using QLoRA~\cite{dettmers2023qlora} on the Thai instruction dataset, named Wangchan-FLAN-6M\footnote{\url{https://huggingface.co/datasets/airesearch/WangchanX-FLAN-v6.1}}.
%
We call the resultant model WangchanLION-V3-8B.
%
%
For comparison, we employ Llama 3.1 8B base, SEA-LION-V3-8B-base, and Typhoon 2 8B base with the same SFT setting as our model.
For consistency, all models have 8 billion parameters and use Llama-3.1-base as the original base model.
%

%
%

\subsection{Evaluation Benchmark}

To evaluate our trained model, including both Llama and GPT-2, we evaluate them using two benchmarks:
\begin{compactitem}
    \item Thai LLM Benchmark~\cite{thaillm-leaderboard} is a benchmark suite for Southeast Asian Languages~\cite{lovenia-etal-2024-seacrowd}. It can evaluate LLMs on NLG and NLU tasks. For GPT-2, we use the NLG task only.
    \item SEA-HELM~\cite{susanto2025seahelmsoutheastasianholistic} is an LLM benchmark designed to evaluate SEA languages, including Thai ones. We remove NLR from the benchmark because the benchmark uses a low-quality machine-translated dataset, XNLI~\cite{conneau-etal-2018-xnli}, for which the results from this dataset are unreliable~\cite{agrawal-etal-2024-translation,singh2025globalmmluunderstandingaddressing}. 
\end{compactitem}
These benchmarks allow us to evaluate the base and instruction models.
For the evaluation of GPT-2, we opted not to include metrics related to NLU, MT-bench, and safety. This decision is based on the observation that the scores for GPT-2 were significantly lower than the established baseline, resulting in normalized scores of zero.
The primary reason for this issue lies in the limited capacity of the 124-million parameter GPT-2 model, which is insufficient for effectively processing complex prompts for the NLU tasks. Consequently, our evaluation focused exclusively on the NLG and Instruction-Following Evaluation (IF) tasks.
%






\section{Experimental Results}

\begin{table*}[h!]
\centering
\scalebox{0.8}{
\begin{tabular}{ l  |l  |l  |l  |l  |l  |l  |l  |l }
\hline
 & \multicolumn{3}{|c|}{SEA-HELM} & \multicolumn{5}{|c}{Thai LLM Benchmark} \\
\hline
\diagbox[width=11em]{Training Data}{Task} & NLG & IF & Avg. & ENG->TH & TH->ENG & XLSum & iApp & Avg. \\
\hline
\multicolumn{9}{l}{\textit{Common crawl}}\\
\hline
baseline & 3.09 & 11.00& 7.04 & \textbf{0.13} & 0.17 & 6.98 & 1.05 & 2.08 \\

+ Language Identification & 4.43 & 14.00& 9.21 & 0.08 & 0.21 & 6.92 & 1.08 & 2.07 \\

+ Quality Filters & \textbf{12.29} & 23.00& 17.64 & 0.09 & \textbf{0.24} & 8.11 & \textbf{1.54} & 2.50\\

+ Deduplication & 6.59 & 18.00& 12.29 & 0.09 & 0.21 & 8.37 & 1.43 & 2.52 \\

+ Content Filters & 10.60& \textbf{25.00}& \textbf{17.80} & 0.07 & 0.21 & \textbf{8.66} & 1.17 & \textbf{2.53} \\
\hline
\multicolumn{9}{l}{\textit{FineWeb2}}\\
\hline
FineWeb2 & 9.13 & 13.56& 11.34 & \textbf{0.08} & 0.13 & \textbf{8.71} & 1.28 & \textbf{2.55} \\

FineWeb2 + our pipeline & \textbf{15.37} & \textbf{16.00}& \textbf{15.68} & 0.07 & \textbf{0.18} & 8.33 & \textbf{1.37} & 2.49 \\
\hline

\end{tabular}}
    \caption{The result of SEA-HELM and Thai LLM Benchmark evaluation for GPT-2}
    \label{tab:sea_helm_cc}
    \label{tab:gpt2_nlg_seacrowd}
\end{table*}


\subsection{Data Ablation Study} 






In this study, we want to explore the effectiveness of our data pipeline using GPT-2's downstream task performances as the indicator.
In particular, we apply our pipeline to unclean data, Common Crawl, and cleaned data, FineWeb2.
\emph{Our data cleaning pipeline should yield improvement on all data since we designed it specifically for Thai}.

\subsubsection{Ablation Design}
We describe the setup of the data ablation studies on Common Crawl and FineWeb2 as follows:

\noindent
\textbf{Common Crawl}. We train a GPT-2 model on each of five dataset variations, with each containing 10 billion tokens. 
Each dataset variation builds upon the previous one by incrementally adding a new cleaning component, allowing us to measure each step's impact.
\begin{compactitem}
    \item \textbf{Baseline}: The raw Thai Common Crawl corpus.
    \item \textbf{+ Language Identification}: Baseline with only Thai-language document retained.
    \item \textbf{+ Quality Filters}:  Adds the quality filtering from Section~\ref{sec:quailty_filters}.
    \item \textbf{+ Deduplication}: Adds URL and text deduplication
    \item \textbf{+ Content Filters}: Adds filters for
adult content, gambling content, and PII as we proposed in Section~\ref{sec:content_filters}.
\end{compactitem}

\noindent
\textbf{FineWeb2}.
We aim to assess the effectiveness of our data cleaning pipeline. To do this, we compare the performance of the model trained on a clean baseline dataset with the performance on the same data after our pipeline has further processed it.
We train a separate GPT-2 model on 10B tokens from each of the two versions of the FineWeb2 dataset, as follows:
\begin{compactitem}
    \item \textbf{FineWeb2}: We use the original data from FineWeb 2, where the data is already cleaned using their pipeline. 
    \item \textbf{FineWeb2 + our pipeline}: We perform the second cleaning with our data cleaning pipeline.
\end{compactitem}

\subsubsection{Before and after applying our pipeline.}
As shown in Table~\ref{tab:size_step}, we found that our pipeline can clean and reduce the size of the dataset. 
For Common Crawl, we can reduce the dataset from 202 million documents to 25.1 million, resulting in a cleaner dataset than the raw source.
Similarly, we also reduce the size of FineWeb2 by half.
We found that most of the text was removed by our C4, Gopher, and gambling-related content filters. These samples were low-quality and did not affect downstream task performance.
In the following, we will discuss the effect of our pipeline when applied to both datasets in downstream tasks.




\subsubsection{Using our data cleaning pipeline with unclean data (Common Crawl)}

Table~\ref{tab:size_step} shows the results of each filtering step: selecting for Thai-language documents yielded about 27 million documents, applying quality filters affected 139.4 million, deduplicating based on URL and text overlap affected 9.6 million, and filtering for inappropriate content (e.g., adult material, gambling, PII) affected 0.9 million.


As shown in Table~\ref{tab:sea_helm_cc}, the Common Crawl data that has gone through all of our cleaning components yields the best results on both SEA-HELM and the Thai LLM benchmark.
%
%
%
When incrementally adding a new component, we generally observe an improvement on both benchmarks, except for the deduplication process, where we see a drop in the average SEA-HELM score. However, the score increases and surpasses the dip after adding the content filters, yielding the best performance.
This implies that our cleaning pipeline can filter out low-quality data from an uncleaned dataset, resulting in better performance in downstream tasks.

\begin{table}[h!]
\centering
\scalebox{0.8}{
\begin{tabular}{ l | l | l }
\hline
Dataset & Common Crawl & FineWeb2 \\
\hline
Baseline & 231Bt/202Md & 51.4Bt/35.9Md \\

Language Identification & 206Bt/175Md & 51.3Bt/35.9Md \\

Quality Filters & 54.6Bt/35.6Md & 30.4Bt/19.1Md \\

Deduplication & 40.7Bt/26.0Md & 27.8Bt/17.9Md \\

Content Filters & 38.6Bt/25.1Md & 26.0Bt/17.1Md \\
\hline

\end{tabular}}
    \caption{The result of the dataset size in each step from our pipeline (Bt is billions of tokens and Md is millions of documents)}
    \label{tab:size_step}
\end{table}

\subsubsection{Using our data cleaning pipeline with cleaned data (FineWeb2)}
%
%
Next, we question our data pipeline: \emph{can our data cleaning pipeline improve the data that has already been cleaned, like FineWeb2?}
To answer this question, we compare the original FineWeb2 and FineWeb2 processed with our pipeline.
Our findings indicate that the original FineWeb2 performs slightly better overall than the version further cleaned with our pipeline in the Thai LLM Benchmark evaluation, as shown in Table~\ref{tab:gpt2_nlg_seacrowd}.
FineWeb2 processed with our pipeline substantially outperforms the original FineWeb2 on average in the SEA-HELM evaluation, as shown in Table~\ref{tab:sea_helm_cc}.
Moreover, we also found that, before applying our data pipeline, FineWeb2 had unwanted texts, e.g., low-quality, duplication of URLs and text overlap, adult content, gambling content, and PII. 
Further processed with our pipeline, we can significantly reduce its size from 35.9 million documents to 17.1 million, as shown in Table~\ref{tab:size_step}.
This result is achieved by further cleaning and filtering data from FineWeb2. 
This implies that our benchmark can improve data quality by filtering low-quality data from the cleaned data.
In addition, this also implies that the original cleaning data pipeline from FineWeb2 might not be applicable or suitable for Thai texts.

\subsection{Downstream task results}
We assess the effectiveness of our training data by comparing the model trained on our training data with models trained on other pre-training corpora.
%
We hypothesize that the model trained on cleaner and better data quality should yield more improvement than the others.

\begin{figure}[h]
  \centering
  \includegraphics[width=0.5\textwidth]{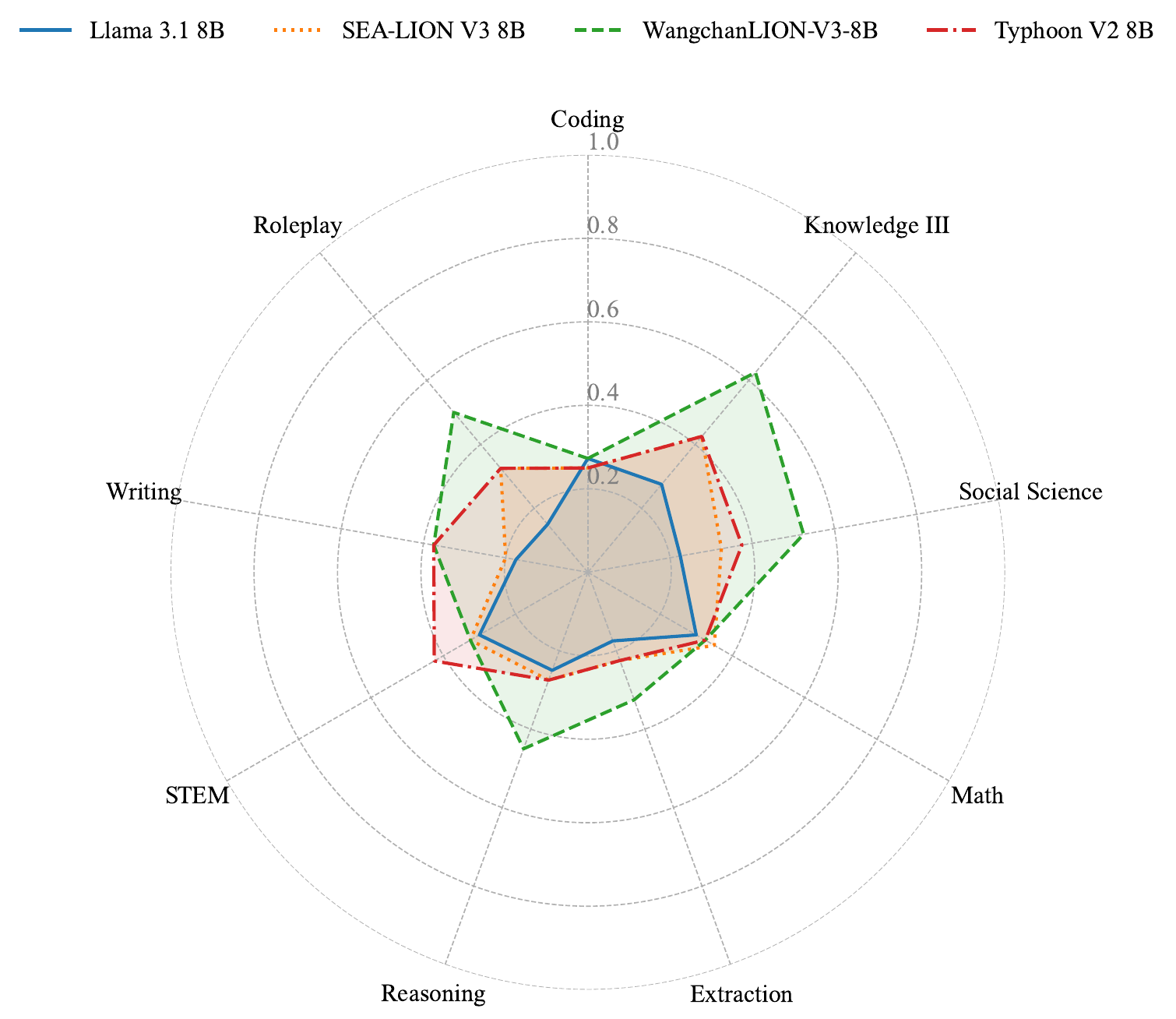}
    \caption{
    The results from the Thai MT-Bench using four instruction models trained on the same data.
    }
  \label{fig:three_figures-mt-beach-cpt}
\end{figure}

\noindent
\textbf{SEA-HELM}. As shown in Table~\ref{tab:seahelm_results}, we evaluate all the base models using the same SFT data, as mentioned in Section~\ref{sec:cpt_setting}. 
Our model achieves the best overall performance among the other three models.
In particular, we outperform our base model, SEA-LION V3 8B, in most cases. 
Furthermore, we perform a more detailed analysis on the chat benchmark, MT-Bench, in Figure~\ref{fig:three_figures-mt-beach-cpt}.
We found that, when using our base model, the performance of the Knowledge III (cultural evaluation) category is higher than that of other models.
This emphasizes the importance of using our data, which can also yield improvement in terms of Thai cultural knowledge.
In addition, we have also gained improvements in the Roleplay and Reasoning categories.

\begin{table}[h!]
\centering
\scalebox{0.6}{
\begin{tabular}{ l | l | l | l | l }
\hline
Task & \rotatebox[origin=c]{90}{Llama 3.1 8B} & \rotatebox[origin=c]{90}{SEA-LION V3 8B} & \rotatebox[origin=c]{90}{Typhoon 2 8B} & \rotatebox[origin=c]{90}{\cptname} \\
\hline
NLU & 46.59 & 54.35 & \underline{60.55} & \textbf{62.33} \\
Safety & 2.47 & \underline{17.21} & \textbf{23.74} & 6.62 \\
NLG & \underline{54.09} & 51.01 & 51.69 & \textbf{54.79} \\
IF & 31.00 & \underline{44.00} & 38.00 & \textbf{52.00} \\
MT-Bench & 23.58 & 30.00 & \underline{33.33} & \textbf{41.36} \\
Avg. & 31.54 & 39.31 & \underline{41.46} & \textbf{43.42} \\
\hline
\end{tabular}}
\caption{The performance comparison in SEA-HELM between existing base models and \cptname  model when utilizing the same SFT dataset.}
\label{tab:seahelm_results}
\label{tab:all_base}
\end{table}




\noindent
\textbf{Thai LLM Leaderboard}. We also conducted an experiment using the benchmark that was formulated focusing on Thai contexts, namely the Thai LLM Leaderboard.
As shown in Table \ref{tab:thaillm_cpt}, we found that our model outperforms other models on the NLG task, but performs lower than other models on the NLU task.
This is because we added extensive Thai pre-training corpora, yielding more Thai fluency and knowledge, resulting in the improvement of the NLG task.
In contrast, the world knowledge will also disappear from our model, resulting in lower performance in NLU (the majority of NLU datasets are not created in Thai contexts, unlike the NLG datasets).
This is consistent with the result from the chat dataset, MT-Bench, that \cptname model yields more fluency in Thai but fails in the world-knowledge subset (STEM).
\begin{table}[htbp]
\centering
\scalebox{0.65}{
\begin{tabular}{l|l |l |l |l |l }\hline
Task&subset& \rotatebox[origin=c]{90}{Llama 3.1 8B} & \rotatebox[origin=c]{90}{SEA-LION V3 8B} & \rotatebox[origin=c]{90}{Typhoon 2 8B} & \rotatebox[origin=c]{90}{\cptname} \\\hline
 NLU&Wisesight & 34.82 & 52.30 & \textbf{62.11} & \underline{52.64}\\
 &XCOPA & 69.40& \underline{78.20}& \textbf{81.00}& 71.40 \\
 &Belebele & 54.67 & \underline{63.44}& \textbf{65.44} & 58.56 \\
 &Avg.& 52.96& \underline{64.64}& \textbf{69.51}& 60.86 \\ \hline
 NLG&XLSum & 48.18 & 48.88 & \underline{54.33}& \textbf{57.01} \\
 &iApp& 74.45 & 79.81 & \underline{80.15}& \textbf{80.99}\\
 &ENG->TH & 21.15 & 21.54 & \underline{26.09}& \textbf{29.00}\\
 &TH->ENG & 49.23 & 51.20& \underline{51.94}& \textbf{52.34} \\
 &Avg.& 48.25 & 50.36 & \underline{53.13}& \textbf{54.84}\\ \hline
\end{tabular}}
\caption{The performance comparison in Thai LLM Leaderboard between existing base models and \cptname model when utilizing the same SFT dataset.}
\label{tab:thaillm_cpt}
\end{table}

\section{Conclusion and Outlook}

We propose Mangosteen, an openly released Thai pre‑training data pipeline and corpus that closes the transparency gap in Thai continual pre-training (CPT) models. 
Our Thai‑specific filters shrink raw Common Crawl data from 202 million to 25 million documents and raise SEA-HELM NLG from about 3 to roughly 11 points. 
An 8 billion-parameter model trained on the resulting 47 billion-token corpus surpasses SEA-LION-v3 and Llama-3.1 on Thai benchmarks by about two points. 
To ensure full reproducibility, we release the pipeline code, the cleaned corpus, and all CPT and SFT checkpoints.

\paragraph{Open-data expansion.}
Mangosteen currently includes only public‑domain or explicitly Creative Commons licensed text. 
We chose this approach to respect the legal rights of content owners. 
We are working with agencies and rights-holders to unlock more than 10 billion additional tokens under permissive licenses.

\paragraph{Pre-training Cost and Shared Knowledge.}
Computational expense is still the main obstacle to studying the impact of data pipeline design. 
For this project, we secured 2,000 GPU‑hours on H100 hardware, giving us effectively one full trial. 
Until reliable low‑compute pre‑training methods emerge, maximizing the public value of every costly run through complete knowledge sharing is essential to the collective progress of this field.
With this limited experimentation budget, our model attains strong overall scores (approximately 2 points above recent Thai baselines), confirming the effectiveness of Thai‑aware curation. 
The central contribution, however, is not the lead itself but the openly released pipeline, corpus, and checkpoints that let others build on and quickly surpass these results.

\subsection*{Acknowledgement}

This research is supported by the National Research Foundation, Singapore, under its National Large Language Models Funding Initiative. Any opinions, findings and conclusions or recommendations expressed in this material are those of the author(s) and do not reflect the views of National Research Foundation, Singapore.


\bibliography{tacl2021}
\bibliographystyle{acl_natbib}

\iftaclpubformat

\fi

\appendix

\onecolumn

\section{GPT-2 Config}
\label{appendix:gpt2_config}
We use the GPT-2 config as follows: sequence length: 1024, n\_layer: 12, n\_head: 12, n\_embd: 768, vocab\_size: 50257, learning rate: 0.0006, num\_train\_epochs: 1

\section{Language Identification}
\label{appendix:lang_ident}

In this section, we compare three language identification libraries—langdetect, lingua, and fastText—and present our findings. The Dolma codebase supports five language identifiers: cld3, pycld2, langdetect, Lingua, and fastText. While fastText was used in the original Dolma implementation, its effectiveness on Thai texts remains uncertain. Thus, our evaluation focuses on langdetect, lingua, and fastText, excluding cld3 and pycld2 due to their outdated models and limited adoption.

We conducted our experiments on a subsample of the raw Thai Common Crawl dataset, consisting of the first 200,000 documents. Each of the three language identifiers was run separately on this subsample to generate a language confidence score for each document. We used a threshold of 0.5, the same cut-off used in the Dolma English corpus, to determine language inclusion. Based on this threshold, the number of documents retained by each language identifier is shown in Table~\ref{tab:language_identifier_counts}.

\begin{table}[h!]
\centering
\begin{tabular}{|l|r|}
\hline
\textbf{Language Identifier} & \textbf{Documents Remaining} \\
\hline
Lingua & 177,221 \\
\hline
Langdetect & 181,872 \\
\hline
FastText & 187,772 \\
\hline
\end{tabular}
\caption{Number of Documents Remaining After Applying a 0.5 Threshold}
\label{tab:language_identifier_counts}
\end{table}

In our analysis, we observed that lingua excluded the highest number of documents compared to the other two language identifiers. Regarding processing time, langdetect exhibited the slowest performance, taking approximately 40 minutes to process all 200,000 documents. In contrast, both lingua and fastText completed the processing in roughly 3 minutes.
Subsequently, we calculated the proportions of Thai characters, vowels, and intonation marks in the documents retained by each language identifier and compared their distributions.

\begin{figure}
    \centering
    \includegraphics[width=0.75\linewidth]{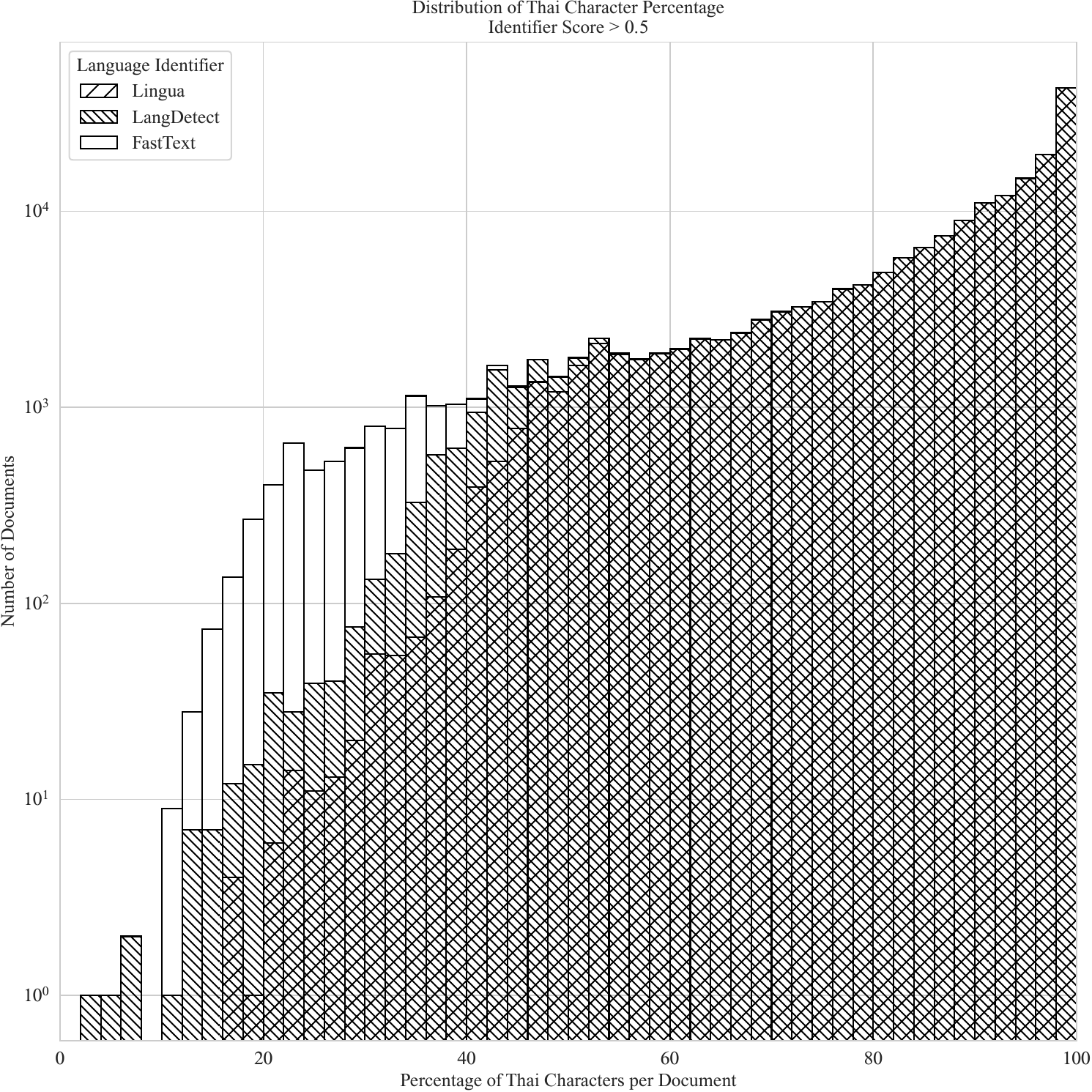}
    \caption{Comparison of the distribution of Thai character ratios for three language identifiers: Lingua, Langdetect, and FastText. Notably, FastText assigns a language score $\geq 50$ to a portion of documents where the Thai character ratio is only between 10 and 40, indicating potential misclassification.}
    \label{fig:thai_char_prop_dis_0.5}
\end{figure}

As illustrated in Figure~\ref{fig:thai_char_prop_dis_0.5}, the majority of documents exhibit a high proportion of Thai characters, vowels, and intonation marks, indicating that the language identifiers are generally effective in detecting Thai text. However, it is evident that \texttt{fastText} flags a larger number of documents as Thai, even when those documents contain a relatively low proportion of Thai-specific characters—specifically, less than 40\%. This suggests that \texttt{fastText} may be more permissive or less precise in its classification compared to \texttt{Lingua} and \texttt{Langdetect}.

To further evaluate \texttt{fastText}'s performance, we isolated the subset of documents in which the combined proportion of Thai characters, vowels, and intonation marks was less than or equal to 40\%. We then analyzed the distribution of language confidence scores assigned by \texttt{fastText} to these documents. This analysis was intended to assess the reliability of its language identification in cases where Thai-specific character usage is minimal.

\begin{figure}[!tbp]
  \centering
  \begin{minipage}[b]{0.45\textwidth}
    \includegraphics[width=\textwidth]{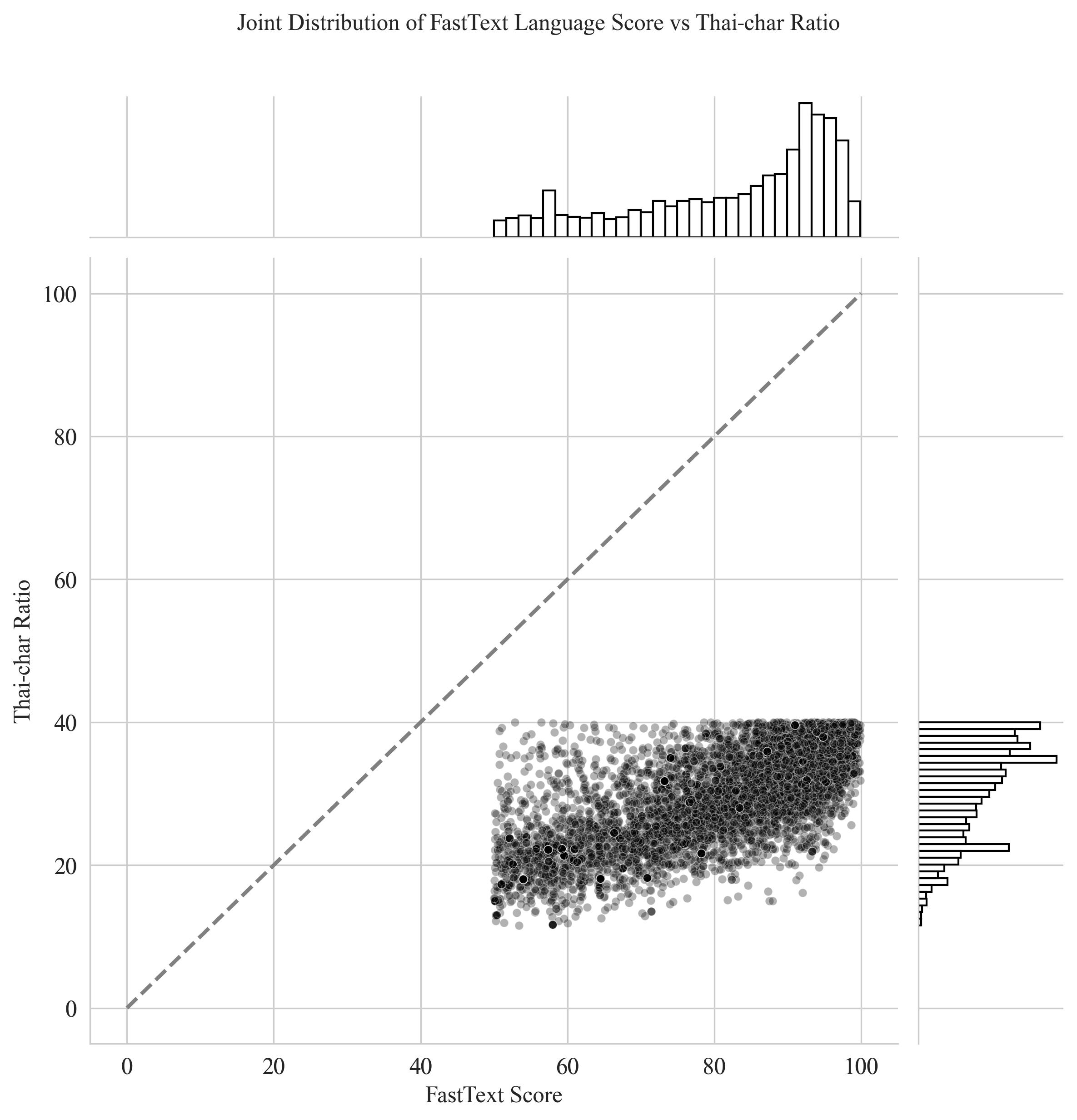}
    \caption{Scatter plot showing that even when the \texttt{fastText} score is high, the proportion of Thai characters can be very low. The overall trend deviates from what would be expected in an ideal language classification scenario.}
    \label{fig:Fasttext lang score vs Thai char ratio 1}
  \end{minipage}
  \hfill
  \begin{minipage}[b]{0.45\textwidth}
    \includegraphics[width=\textwidth]{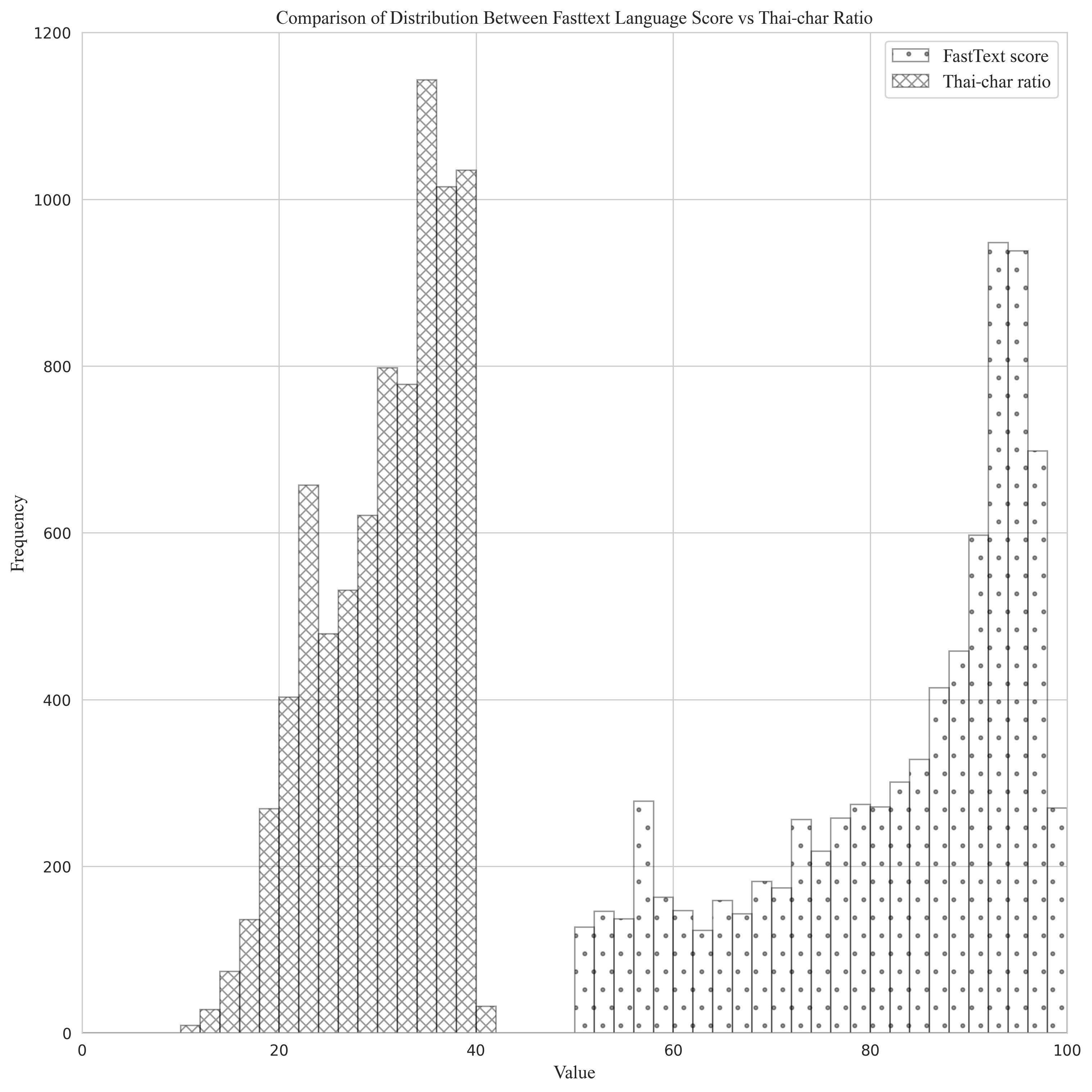}
    \caption{Comparison between Thai character ratio and \texttt{fastText} language scores across documents, highlighting inconsistencies in the confidence scores relative to Thai character content.}
    \label{fig:Fasttext lang score vs Thai char ratio 2}
  \end{minipage}
\end{figure}

As illustrated in the Figure~\ref{fig:Fasttext lang score vs Thai char ratio 1}, fastText assigns high language confidence scores—often approaching 100\%—to documents where the proportion of Thai characters, vowels, and intonation marks does not exceed 40\%. This indicates a tendency of fastText to exhibit overconfidence in its language identification, potentially misclassifying documents with minimal Thai-specific character content. Notably, 8,008 documents (accounting for 4\% of the 200,000-document sample) were identified by fastText as Thai with confidence scores above 0.5, despite containing 40\% or fewer Thai-specific characters.

To this end, it might seem that the only language identifier good enough for the Thai language is Lingua. However, considering Thai's distinctive script, we concluded that relying on the Lingua language identifier may not be optimal. Therefore, we developed the \texttt{ThaiCharRatioTagger}, a custom language identifier that calculates the percentage of Thai characters, vowels, and intonation marks within a document. This approach simplifies the codebase and offers a more straightforward alternative to complex machine learning models.

\section{Gopher Rules}
As stated in Section 3.1.3, the original English corpus Dolma relies heavily on the Gopher Rules, as it uses all of them. In this work, we aim to follow the same methodology as the original, while adjusting the thresholds of certain rules to better suit the Thai language.

The original Gopher Rules used in the English corpus Dolma consist of 11 rules:
\begin{itemize}
    \item Fraction of characters in the most common n-gram exceeds a threshold
    \item Fraction of characters in duplicate n-grams exceeds a threshold
    \item Contains fewer than 50 or more than 100K words
    \item Median word length is less than 3 or greater than 10
    \item Fraction of words containing alphabetic characters is less than 0.80
    \item Contains fewer than 2 of a predefined set of required words
    \item Fraction of lines in the document starting with a bullet point exceeds 0.90
    \item Fraction of lines in the document ending with an ellipsis exceeds 0.30
    \item Fraction of lines in the document that are duplicated exceeds 0.30
    \item Fraction of characters in duplicated lines exceeds 0.30
\end{itemize}

We then begin by experimenting with the correlations of each rule. We first use the Dolma pipeline to obtain scores on a 200,000-document subsample. 
Specifically, instead of tokenizing by spaces, we use the ICU tokenizer to obtain a list of tokens for each document. We use the set of required words provided by Stopwords ISO~\footnote{\url{https://github.com/stopwords-iso/stopwords-th}}, which essentially corresponds to a list of Thai stopwords. In addition, we divide each document into sentences using the delimiter \texttt{\textbackslash n+}

\begin{figure}[!tbp]
  \centering
  \begin{minipage}[b]{0.6\textwidth}
    \includegraphics[width=\textwidth]{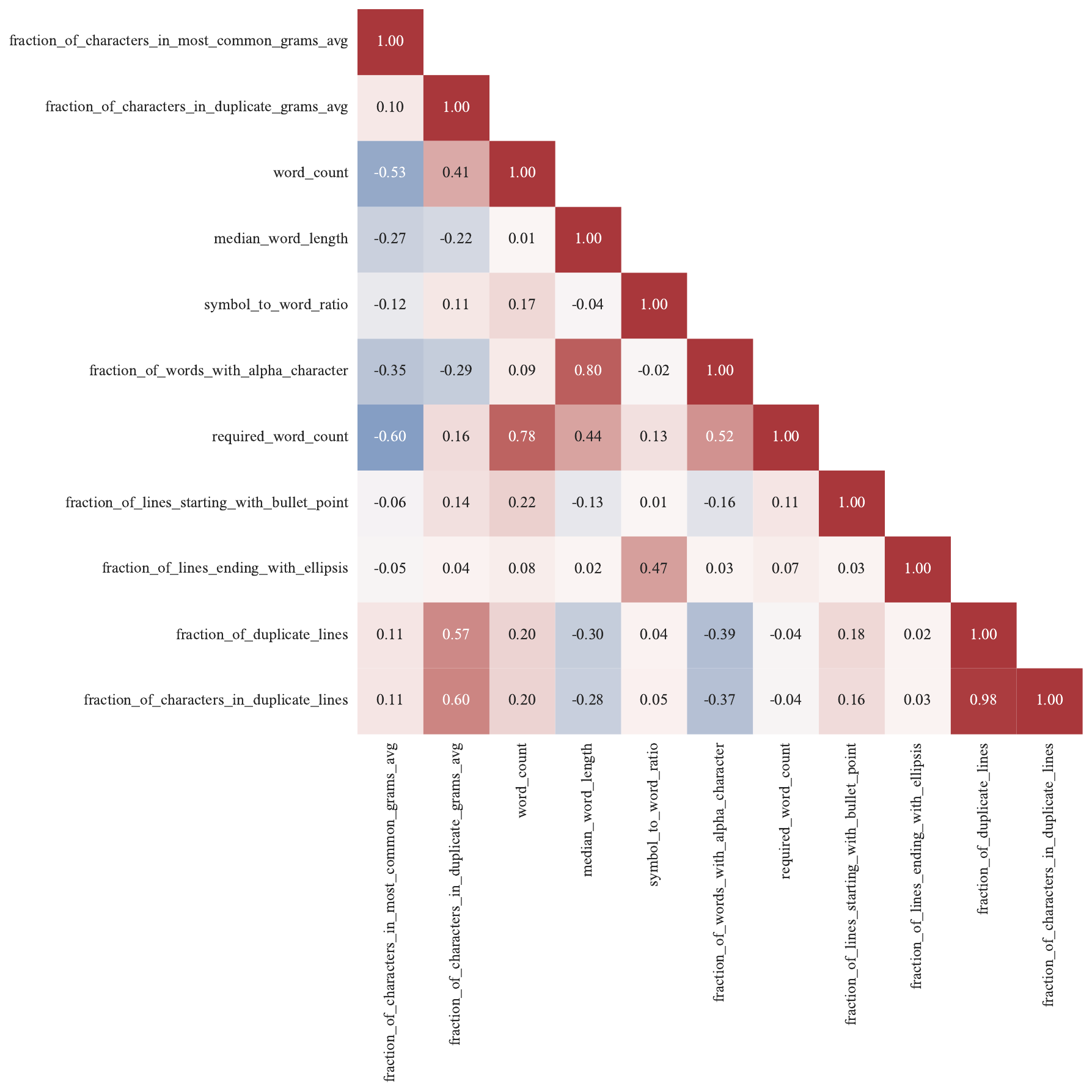}
    \caption{Spearman correlation heatmap of Gopher Rule scores.}
    \label{fig:gopher_correlation_heatmap_spearman}
  \end{minipage}
  \hfill
  \begin{minipage}[b]{0.6\textwidth}
    \includegraphics[width=\textwidth]{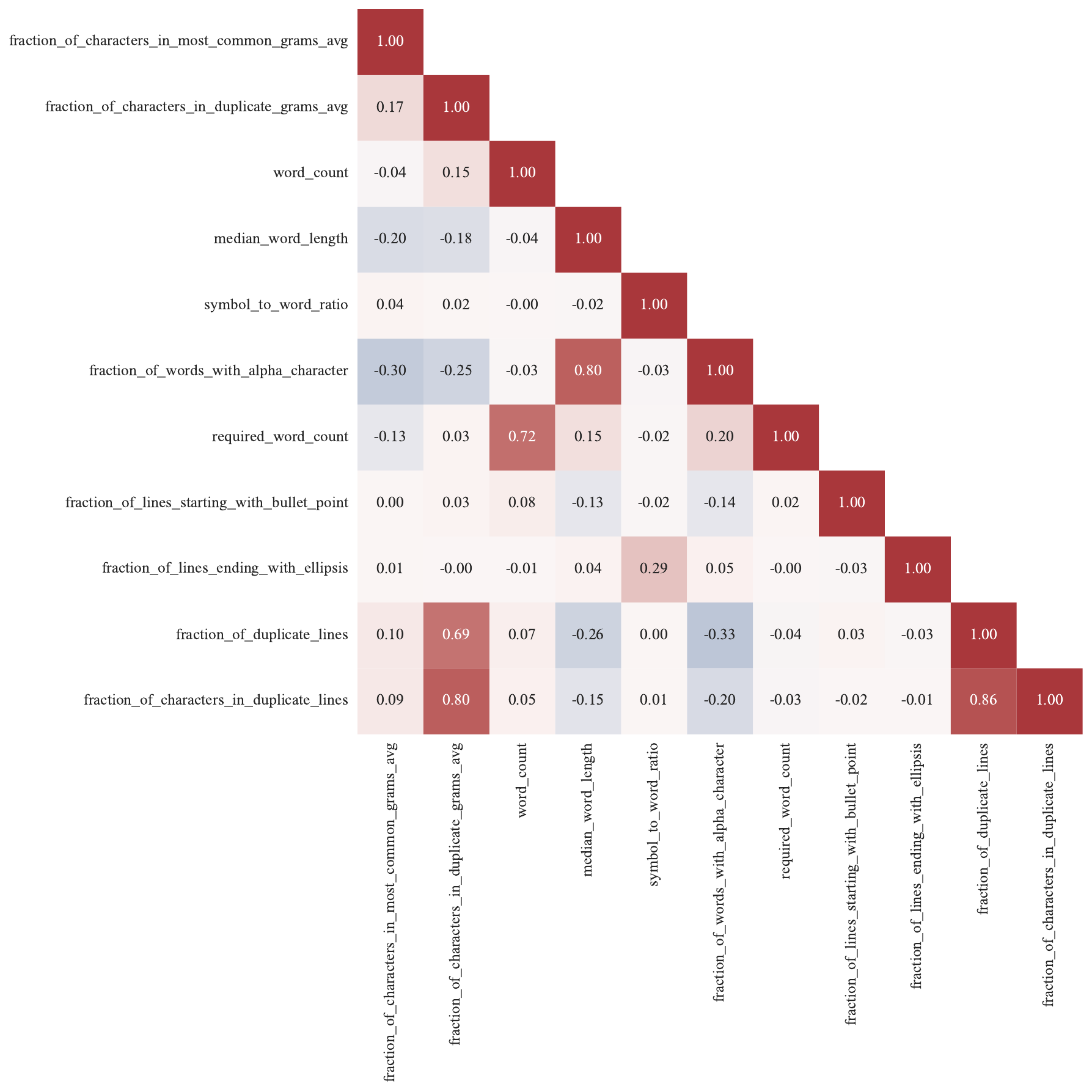}
    \caption{Pearson correlation heatmap of Gopher Rule scores.}
    \label{fig:gopher_correlation_heatmap_pearson}
  \end{minipage}
\end{figure}

In the Dolma work, the authors noted that stacking quality filtering, content filtering, and deduplication results in a positive compounding effect; these rules overlap very little in the texts they remove. Our study comes to a similar conclusion regarding the Gopher rules. As shown in Figure~\ref{fig:gopher_correlation_heatmap_spearman} (Spearman correlation heatmap) and Figure~\ref{fig:gopher_correlation_heatmap_pearson} (Pearson correlation heatmap), we observe only low to moderate correlations between most rules. Some exceptions exist; for example, the "median word length" rule and the "fraction of words with alpha characters" rule are somewhat correlated. This is expected, as a higher median word length generally implies a higher number of alphabetic characters. A similar relationship is observed between the "word count" rule and the "required word count" rule. Based on this analysis, we choose to individually adjust the threshold for each rule. Note that for the rules "fraction of characters in most common n-gram" and "fraction of characters in duplicate n-grams," we use the average value within each group in our analysis, as similar rules tend to exhibit high intra-group correlation.

The next question we address is: Which rules should be adjusted? Should we modify all of them or only a subset? To explore this, we examine the top 10 combinations of rules that remove the largest number of documents.

\begin{table}
\centering
\begin{tabular}{|p{7cm}|l|l|l|}
\hline
Combination of Rules & Count & Percentage & Cumulative Percentage \\
\hline
fraction of words with alpha character & 50,712 & 29.3318 & 29.3318 \\
\hline
fraction of characters in duplicate n grams, fraction of words with alpha character & 17,823 & 10.3088 & 39.6406 \\
\hline
median word length, fraction of words with alpha character & 17,028 & 9.8490 & 49.4896 \\
\hline
fraction of characters in duplicate n grams, median word length, fraction of words with alpha character & 12,812 & 7.4104 & 56.9000 \\
\hline
fraction of characters in duplicate n grams, median word length, fraction of words with alpha character, fraction of duplicate lines, fraction of characters in duplicate lines & 9,103 & 5.2652 & 62.1652 \\
\hline
fraction of characters in duplicate n grams, fraction of words with alpha character, fraction of duplicate lines, fraction of characters in duplicate lines & 8,447 & 4.8857 & 67.0509 \\
\hline
fraction of characters in duplicate n grams, median word length, fraction of words with alpha character, fraction of duplicate lines & 7,631 & 4.4138 & 71.4647 \\
\hline
median word length, fraction of words with alpha character, required word count & 5,268 & 3.0470 & 74.5117 \\
\hline
fraction of characters in duplicate n grams, fraction of words with alpha character, fraction of duplicate lines & 3,694 & 2.1366 & 76.6483 \\
\hline
median word length, fraction of words with alpha character, fraction of duplicate lines & 2,772 & 1.6033 & 78.2516 \\
\hline
\end{tabular}
\caption{Top combinations of Gopher Rules by number of filtered documents}
\label{tab:gopher_rule_combinations}
\end{table}

As shown in Table~\ref{tab:gopher_rule_combinations}, approximately 78\% of the documents are filtered out by these rules or combinations of rules. The rules frequently appearing in this table include: fraction of words with alpha characters, fraction of characters in duplicate n-grams, median word length, fraction of duplicate lines, and required word count. Among these, we choose to further investigate the "median word length" rule, as the other rules can be applied directly to Thai without modification. Although not included in the table, we also specifically examine the "word count" rule, as we believe that the typical word count differs significantly between Thai and English documents. Furthermore, we observe that low word count documents in Thai often correspond to low-quality text data.
Some rules can be modified with minimal data exploration:
\begin{itemize}
    \item The first is the rule "fraction of words with alpha characters." We modify it to "fraction of words with Thai letters" by counting Thai characters instead of alphabetic characters.
    \item For the rule "fraction of lines in document ending with ellipsis," we detect the presence of three dots (...) at the end of each line, as this pattern appears frequently in our corpus.
\end{itemize}

\subsection{Word Count}
To begin our experiment, we use the same dataset described in Appendix~E—the 200,000-document subsample. In the following, we present descriptive statistics of the word counts in this dataset, where each document is tokenized using the ICU tokenizer.

\begin{table}
\centering
\begin{tabular}{| r | r |}
\hline
Statistic & Value \\
\hline
Count & 200{,}000 \\
\hline
Mean & 675.47 \\
\hline
Standard Deviation & 1,964.60 \\
\hline
Minimum & 0 \\
\hline
10th Percentile & 56 \\
\hline
30th Percentile & 131 \\
\hline
50th Percentile (Median) & 283 \\
\hline
70th Percentile & 600 \\
\hline
90th Percentile & 1,556 \\
\hline
95th Percentile & 2,384 \\
\hline
99th Percentile & 4,823.01 \\
\hline
Maximum & 195{,}713 \\
\hline
\end{tabular}
\caption{Descriptive statistics of word counts in the sub-document dataset, including the mean, standard deviation, median, minimum, maximum, and selected percentiles.}
\label{tab:word_count_statistic}
\end{table}

\begin{figure}
    \centering
    \includegraphics[width=1\linewidth]{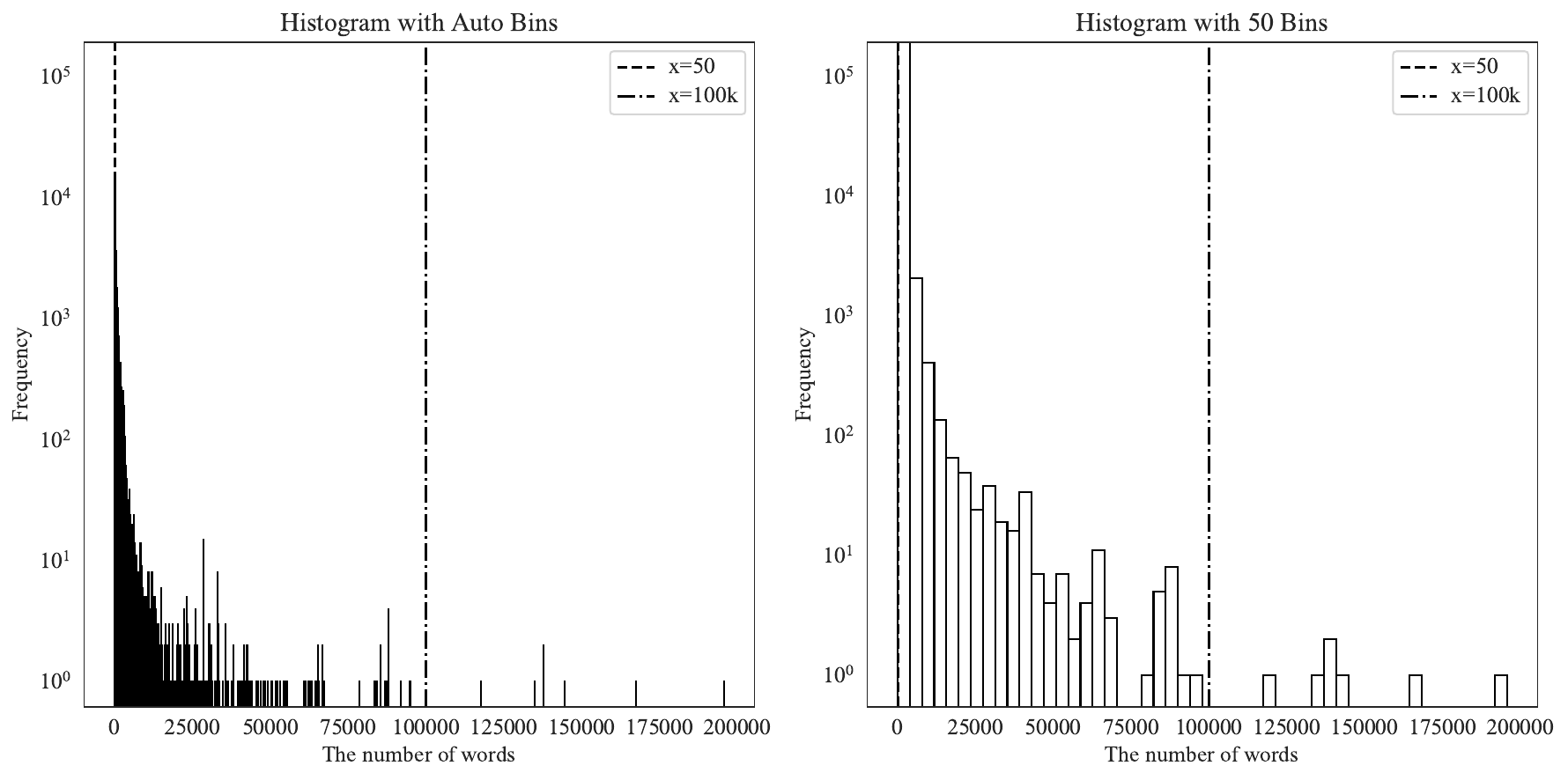}
    \caption{Histograms of word counts in the sub-document dataset. The left plot shows the distribution using automatic binning for finer granularity, while the right plot uses a fixed bin size of 50 to provide a more uniform view.}
    \label{fig:word_count_histogram}
\end{figure}

The distribution of word count is clearly right-skewed. If we apply the Gopher rule cutoffs of 50 and 100,000 words, very few documents are filtered out, and we suspect that many low-quality documents would remain. Upon closer inspection through data binning, we find that the interval containing a high proportion of documents ranges from 21 to 170 words, approximately 35\% of the subsample. We suspect that this interval may contain a disproportionately large amount of low-quality data.

\begin{table}
\centering
\begin{tabular}{| r | r | r | r | r |}
\hline
bin & frequency & cumulative frequency & proportion & cumulative proportion \\
\hline
[ 0. 21.3056] & 2,762 & 2,762 & 1.3810 & 1.3810 \\
\hline
[21.3056 42.6111] & 11,137 & 13,899 & 5.5685 & 6.9495 \\
\hline
[42.6111 63.9167] & 10,778 & 24,677 & 5.3890 & 12.3385 \\
\hline
[63.9167 85.2223] & 14,169 & 38,846 & 7.0845 & 19.4230 \\
\hline
[ 85.2223 106.5279] & 11,247 & 50,093 & 5.6235 & 25.0465 \\
\hline
[106.5279 127.8334] & 8,541 & 58,634 & 4.2705 & 29.3170 \\
\hline
[127.8334 149.139 ] & 7,834 & 66,468 & 3.9170 & 33.2340 \\
\hline
[149.139 170.4446] & 6,929 & 73,397 & 3.4645 & 36.6985 \\
\hline
[170.4446 191.7502] & 5,854 & 79,251 & 2.9270 & 39.6255 \\
\hline
[191.7502 213.0557] & 5,426 & 84,677 & 2.7130 & 42.3385 \\
\hline
[213.0557 234.3613] & 5,083 & 89,760 & 2.5415 & 44.8800 \\
\hline
[234.3613 255.6669] & 4,680 & 94,440 & 2.3400 & 47.2200 \\
\hline
[255.6669 276.9725] & 4,286 & 98,726 & 2.1430 & 49.3630 \\
\hline
[276.9725 298.278 ] & 4,169 & 102,895 & 2.0845 & 51.4475 \\
\hline
[298.278 319.5836] & 3,622 & 106,517 & 1.8110 & 53.2585 \\
\hline
[319.5836 340.8892] & 3,400 & 109,917 & 1.7000 & 54.9585 \\
\hline
[340.8892 362.1948] & 3,639 & 113,556 & 1.8195 & 56.7780 \\
\hline
[362.1948 383.5003] & 3,117 & 116,673 & 1.5585 & 58.3365 \\
\hline
[383.5003 404.8059] & 2,939 & 119,612 & 1.4695 & 59.8060 \\
\hline
[404.8059 426.1115] & 2,598 & 122,210 & 1.2990 & 61.1050 \\
\hline
[426.1115 447.417 ] & 2,432 & 124,642 & 1.2160 & 62.3210 \\
\hline
[447.417 468.7226] & 2,186 & 126,828 & 1.0930 & 63.4140 \\
\hline
[468.7226 490.0282] & 2,288 & 129,116 & 1.1440 & 64.5580 \\
\hline
[490.0282 511.3338] & 2,114 & 131,230 & 1.0570 & 65.6150 \\
\hline
[511.3338 532.6393] & 2,169 & 133,399 & 1.0845 & 66.6995 \\
\hline
[532.6393 553.9449] & 2,082 & 135,481 & 1.0410 & 67.7405 \\
\hline
[553.9449 575.2505] & 2,034 & 137,515 & 1.0170 & 68.7575 \\
\hline
[575.2505 596.5561] & 2,113 & 139,628 & 1.0565 & 69.8140 \\
\hline
[596.5561 617.8616] & 1,843 & 141,471 & 0.9215 & 70.7355 \\
\hline
[617.8616 639.1672] & 1,803 & 143,274 & 0.9015 & 71.6370 \\
\hline
\end{tabular}
\caption{Binning analysis of word counts in the sub-document dataset. A notably high proportion of documents fall within the 21–170 word range. Only the first 30 bins are shown here for display purposes.}
\label{tab:word_count_head_30}
\end{table}

To confirm our assumption, we conduct an experiment using perplexity scores to assess data quality. The model used for this experiment is \texttt{airesearch/wangchanbart-base}. The results are presented in the table below.

\begin{table}
\centering
\begin{tabular}{| r | r | r | r | r |}
\hline
\textbf{Statistic} & \textbf{Baseline} & \textbf{Gopher Rules} & \textbf{Experiment} & \textbf{Short Text} \\
\hline
Count & 200,000 & 182,976 & 126,596 & 70,635 \\
\hline
Mean & 1143.80 & 1101.69 & 1074.90 & 1137.07 \\
\hline
Standard Deviation & 4414.26 & 2162.41 & 2359.05 & 1583.46 \\
\hline
Minimum & 1.37 & 1.37 & 1.37 & 5.37 \\
\hline
10th Percentile & 176.05 & 171.94 & 169.55 & 187.32 \\
\hline
30th Percentile & 402.48 & 397.45 & 390.92 & 416.09 \\
\hline
50th Percentile (Median) & 710.95 & 706.76 & 693.37 & 724.38 \\
\hline
70th Percentile & 1136.51 & 1126.69 & 1081.74 & 1240.15 \\
\hline
90th Percentile & 2163.38 & 2135.86 & 1985.92 & 2372.85 \\
\hline
95th Percentile & 3153.02 & 3096.17 & 2922.14 & 3298.03 \\
\hline
99th Percentile & 7660.35 & 7435.88 & 8193.99 & 5539.36 \\
\hline
Maximum & 814{,}671.56 & 539{,}081.63 & 539{,}081.63 & 51{,}177.91 \\
\hline
\end{tabular}
\caption{Perplexity statistics across four experimental settings. Applying the Gopher Rules results in a noticeable decrease in perplexity scores, with a further reduction observed when increasing the lower bound in our experiment. The "Short Text" setting is included to illustrate that, despite comprising 38\% of the data, it yields perplexity scores nearly as high as the baseline, indicating lower quality.}
\label{tab:word_count_perplexity}
\end{table}

\begin{figure}
    \centering
    \includegraphics[width=1\linewidth]{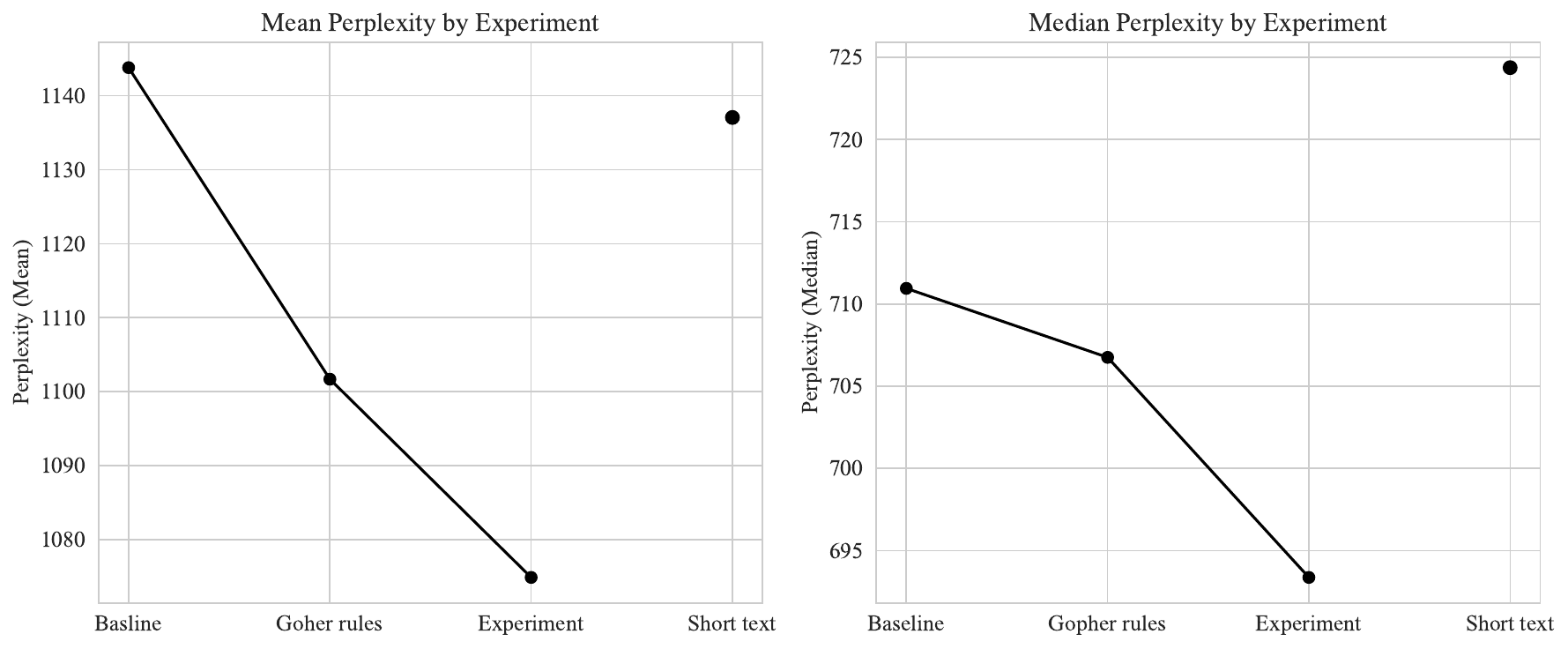}
    \caption{Trends in the mean and median perplexity scores across different experimental settings}
    \label{fig:word_count_perplexity_line}
\end{figure}

\begin{table}
\centering
\begin{tabular}{| c | c | c |}
\hline
Settings & H-statistics & P-value \\
\hline
Baseline vs Experiment & 135.063 & 0.00 \\
\hline
Gopher rules vs Experiment & 73.337 & 0.00 \\
\hline
Baseline vs Gopher rules vs Experiment & 138.922 & 0.00 \\
\hline
\end{tabular}
\caption{Kruskal–Wallis H-test results for perplexity scores across different experimental settings. The $p$-values for all comparisons are 0.00, indicating statistically significant differences in the population medians and rejecting the null hypothesis that the medians are equal.}
\label{tab:word_count_kruskal_text}
\end{table}

Table~\ref{tab:word_count_perplexity} presents the perplexity statistics under four different settings. In the \textbf{Baseline} setting, no filtering is applied; the perplexity scores represent the entire 200{,}000-document subsample. In the \textbf{Gopher Rules} setting, documents are filtered using the original Gopher cutoffs of 50 and 100{,}000 words. The \textbf{Experiment} setting modifies the lower cutoff, increasing it from 50 to 171, in order to filter out all low word count documents, which we suspect may be of lower quality. Finally, the \textbf{Short Text} setting isolates documents with word counts between 21 and 170. Notably, the perplexity scores in this group are nearly as high as those in the baseline, supporting our hypothesis that short documents in this range are likely to be of lower quality. From these statistics, we observe that the \textbf{Experiment} setting yields the lowest mean and median perplexity scores, suggesting that removing very short documents improves overall data quality. To further support this conclusion, we report the Kruskal–Wallis H-test statistic and corresponding $p$-values in Table~\ref{tab:word_count_kruskal_text}. In all cases, the $p$-values indicate a statistically significant difference, leading us to reject the null hypothesis that the medians across settings are equal.

\begin{figure}
    \centering
    \includegraphics[width=1.0\linewidth]{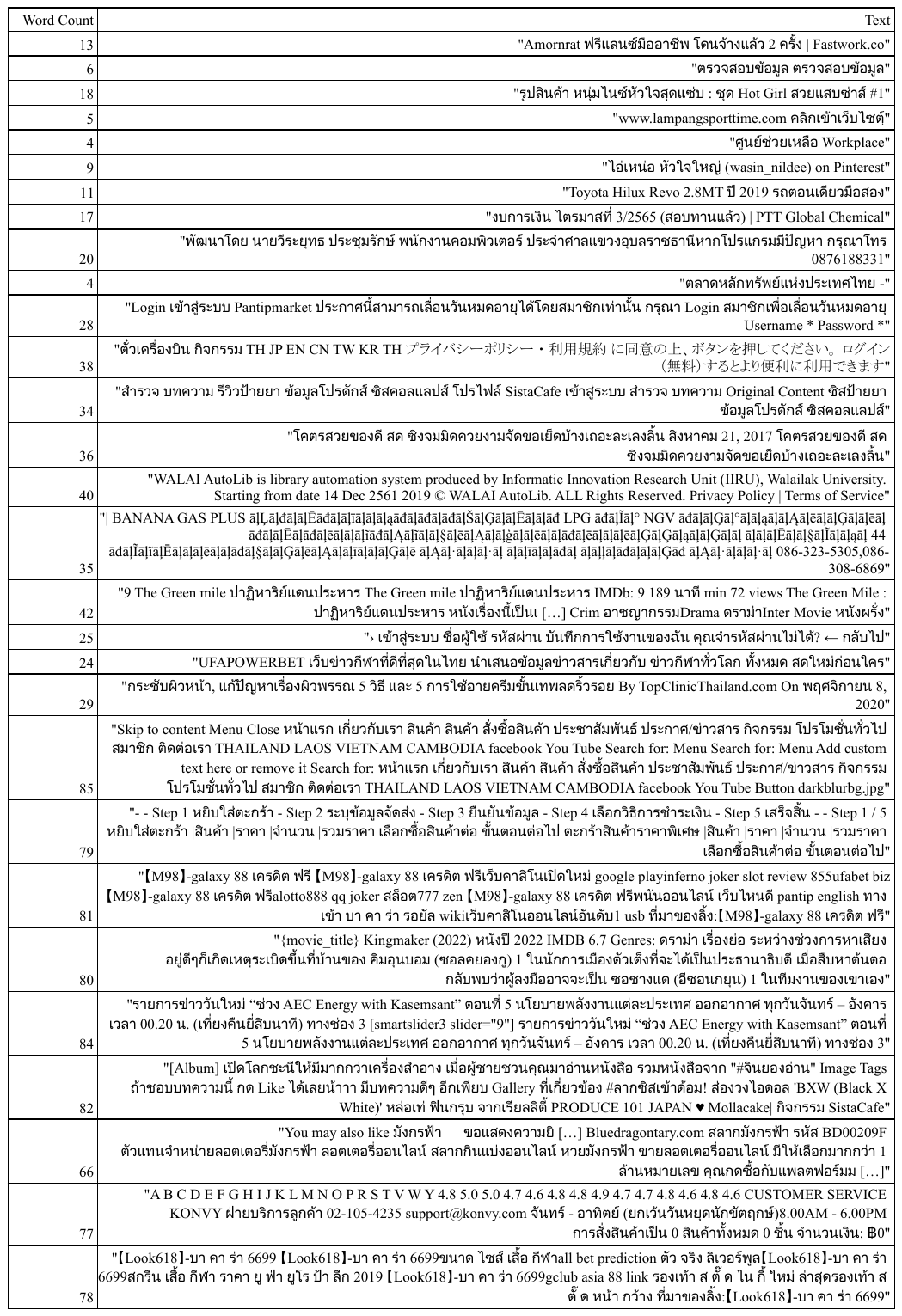}
    \caption{Examples of low word count documents. These examples illustrate the generally low quality of such content. Notably, some documents include material from illegal gambling websites and adult content sources.}
    \label{fig:example_of_low_word_count_documents}
\end{figure}

Based on this experiment, and specifically for the Thai language, we conclude that the upper bound of the word count rule can remain unchanged, while the lower bound should be increased. In our pipeline, we set the lower bound to 200 and retain the original upper bound of 100,000. Ideally, the optimal value for the lower bound should be determined through a series of data ablation experiments. However, due to limited resources and computational constraints, we were unable to perform such extensive evaluations.

\subsection{Median Word Length}

\begin{table}
\centering
\begin{tabular}{| r | r |}
\hline
\textbf{Statistic} & \textbf{Value} \\
\hline
Count & 200{,}000 \\
\hline
Mean & 3.23 \\
\hline
Standard Deviation & 0.72 \\
\hline
Minimum & 0 \\
\hline
10th Percentile & 3 \\
\hline
30th Percentile & 3 \\
\hline
50th Percentile (Median) & 3 \\
\hline
70th Percentile & 3.5 \\
\hline
90th Percentile & 4 \\
\hline
95th Percentile & 4 \\
\hline
99th Percentile & 5 \\
\hline
Maximum & 54 \\
\hline
\end{tabular}
\caption{Descriptive statistics of median word length in the sub-document dataset, including the mean, standard deviation, median, minimum, maximum, and selected percentiles.}
\label{tab:median_word_legnth_statistics}
\end{table}

\begin{figure}
    \centering
    \includegraphics[width=1\linewidth]{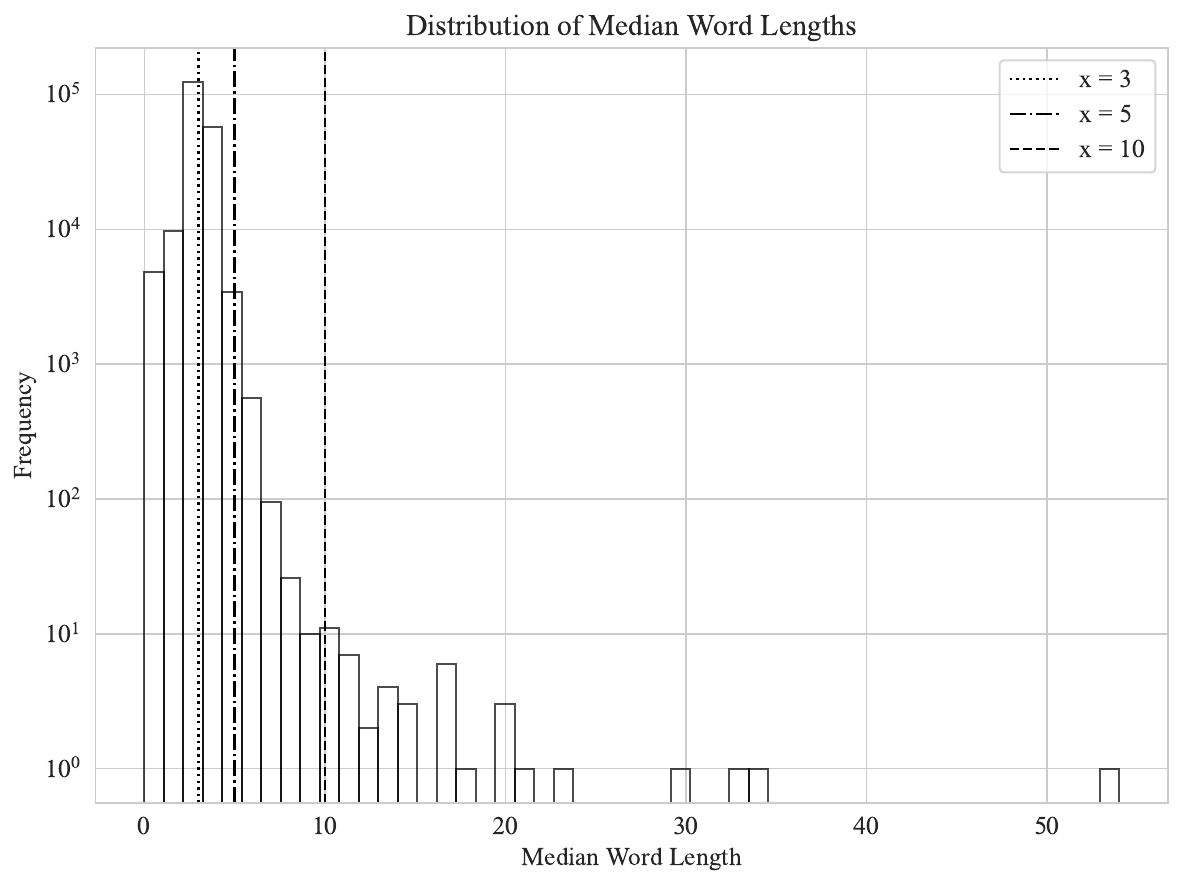}
    \caption{Histogram of the distribution of median word lengths in the sub-document dataset. The distribution is right-skewed. Vertical lines are drawn at $x = 3$ and $x = 10$ to indicate the threshold values used in the original Gopher Rules. An additional line at $x = 5$ marks the 99th percentile in our dataset.}
    \label{fig:median_word_length_histogram}
\end{figure}
\begin{table}
\centering

\begin{tabular}{| l | r | r | r |}
\hline
\textbf{median word length} & \textbf{count} & \textbf{percentage} & \textbf{cumulative percentage} \\
\hline
0 & 4 & 0.0020 & 0.0020 \\
\hline
1 & 4,985 & 2.4925 & 2.4945 \\
\hline
2 & 9,909 & 4.9545 & 7.4490 \\
\hline
3 & 126,156 & 63.0780 & 70.5270 \\
\hline
4 & 55,471 & 27.7355 & 98.2625 \\
\hline
5 & 2,880 & 1.4400 & 99.7025 \\
\hline
6 & 433 & 0.2165 & 99.9190 \\
\hline
7 & 83 & 0.0415 & 99.9605 \\
\hline
8 & 26 & 0.0130 & 99.9735 \\
\hline
9 & 10 & 0.0050 & 99.9785 \\
\hline
10 & 11 & 0.0055 & 99.9840 \\
\hline
11 & 7 & 0.0035 & 99.9875 \\
\hline
12 & 2 & 0.0010 & 99.9885 \\
\hline
13 & 3 & 0.0015 & 99.9900 \\
\hline
14 & 1 & 0.0005 & 99.9905 \\
\hline
15 & 3 & 0.0015 & 99.9920 \\
\hline
17 & 6 & 0.0030 & 99.9950 \\
\hline
18 & 1 & 0.0005 & 99.9955 \\
\hline
20 & 3 & 0.0015 & 99.9970 \\
\hline
21 & 1 & 0.0005 & 99.9975 \\
\hline
23 & 1 & 0.0005 & 99.9980 \\
\hline
30 & 1 & 0.0005 & 99.9985 \\
\hline
33 & 1 & 0.0005 & 99.9990 \\
\hline
34 & 1 & 0.0005 & 99.9995 \\
\hline
54 & 1 & 0.0005 & 100.0000 \\
\hline
\end{tabular}
\caption{Value counts, percentages, and cumulative percentages of median word lengths in the sub-document dataset. The distribution clearly shows that the majority of documents have a median word length of 3 or 4.}
\label{tab:medain_word_lenght_value_counts}
\end{table}

The original Gopher Rules filter out documents with a median word length less than 3 or greater than 10. In our Thai sub-document dataset, we observe that the majority of documents have a median word length between 3 and 4, or more broadly, between 2 and 5. We conducted the same experiment described in Appendix~F.1 to compute perplexity statistics. However, in this case, we modified the upper bound of the median word length rule from 10 to 5, as 5 corresponds to the 99th percentile of our dataset. The results of this experiment are presented below.

\begin{table}
\centering
\begin{tabular}{| r | r | r | r |}
\hline
\textbf{Statistic} & \textbf{Baseline} & \textbf{Gopher Rules} & \textbf{Experiment} \\
\hline
Count & 200{,}000 & 185{,}069 & 184{,}370 \\
\hline
Mean & 1143.80 & 1170.90 & 1168.76 \\
\hline
Standard Deviation & 4414.26 & 2227.70 & 2203.26 \\
\hline
Minimum & 1.37 & 8.71 & 8.71 \\
\hline
10th Percentile & 176.05 & 206.64 & 206.77 \\
\hline
30th Percentile & 402.48 & 434.22 & 434.07 \\
\hline
50th Percentile (Median) & 710.95 & 747.78 & 747.27 \\
\hline
70th Percentile & 1136.51 & 1181.81 & 1181.60 \\
\hline
90th Percentile & 2163.38 & 2238.32 & 2237.23 \\
\hline
95th Percentile & 3153.02 & 3264.85 & 3262.91 \\
\hline
99th Percentile & 7660.35 & 7933.48 & 7917.08 \\
\hline
Maximum & 814{,}671.56 & 191{,}404.94 & 191{,}404.94 \\
\hline
\end{tabular}
\caption{Perplexity statistics across three experimental settings. Applying the Gopher Rules results in a reduction in perplexity scores. However, further adjusting the upper bound of the median word length rule (from 10 to 5) has no significant additional effect on perplexity in our experiment.}
\label{tab:median_word_length_preplexity}
\end{table}

As shown in Tabel~\ref{tab:median_word_length_preplexity}, we found that the perplexity statistics between the two settings did not differ significantly, which suggests that the original Gopher Rule threshold may also be applicable to Thai. Consequently, we decided to retain the original threshold values. The Kruskal–Wallis H-statistic comparing the \textbf{Gopher Rules} and \textbf{Experiment} settings is 0.027, with a $p$-value of 0.869, indicating no statistically significant difference between the medians of the two samples. 

In our Dolma pipeline, we therefore chose not to modify the threshold for this rule. However, we recommend that readers with sufficient computational resources consider conducting a data ablation study to better understand the impact of this rule and identify the optimal threshold for their specific use case.

\subsection{Gopher Rules Changes Summary}

To provide a clear comparison between the original and adapted Gopher Rules used in our work, we present a detailed summary of all modifications in the table below. These changes reflect adjustments made to better accommodate the characteristics of the Thai language.

\begin{table}[ht]
\centering
\begin{tabular}{|p{6cm}|p{7cm}|}
\hline
\textbf{Original Rules} & \textbf{Our Rules} \\
\hline
Fraction of characters in most common n-gram greater than a threshold & Fraction of characters in most common n-gram greater than a threshold \\
\hline
Fraction of characters in duplicate n-grams greater than a threshold & Fraction of characters in duplicate n-grams greater than a threshold \\
\hline
Contains fewer than 50 or more than 100K words & Contains fewer than 200 or more than 100K words \\
\hline
Median word length is less than 3 or greater than 10 & Median word length is less than 3 or greater than 10 \\
\hline
Symbol to word ratio greater than 0.10 & Symbol to word ratio greater than 0.10 \\
\hline
Fraction of words with alpha character less than 0.80 & Fraction of words with Thai letters less than 0.80 \\
\hline
Contains fewer than 2 of a set of required words & Contains fewer than 2 of a set of required words \\
\hline
Fraction of lines in document starting with bullet point greater than 0.90 & Fraction of lines in document starting with bullet point greater than 0.90 \\
\hline
Fraction of lines in document ending with ellipsis greater than 0.30 & Fraction of lines ending with ellipsis or “…” greater than 0.30 \\
\hline
Fraction of lines in document that are duplicated greater than 0.30 & Fraction of lines that are duplicated greater than 0.30 \\
\hline
Fraction of characters in duplicated lines greater than 0.30 & Fraction of characters in duplicated lines greater than 0.30 \\
\hline
– & Contains one of the following: read more \\
\hline
\end{tabular}
\caption{Summary of modifications to Gopher Rules for the Thai language}
\label{tab:summary_of_out_gopher_rules_change}
\end{table}

\section{C4 Rules}

Unlike the Gopher rules, we made minimal modifications to the C4 rules taggers. The original source code for the C4 rules taggers returns the following attributes for a document:

\begin{itemize}
    \item \texttt{has\_curly\_brace}: Indicates the presence of the character \texttt{\{} in the document.
    \item \texttt{has\_lorem\_ipsum}: Checks if the phrase "lorem ipsum" is present in the document.
    \item \texttt{has\_javascript}: Detects the presence of JavaScript code within the document.
    \item \texttt{has\_naughty\_word}: Identifies the presence of inappropriate or profane words.
    \item \texttt{lines\_with\_no\_ending\_punctuation}: Flags lines that do not end with standard punctuation marks.
    \item \texttt{lines\_with\_too\_few\_words}: Flags lines that contain fewer words than a predefined threshold.
    \item \texttt{line\_count}: Records the total number of lines in the document.
\end{itemize}

Firstly, we update the corpus used for detecting inappropriate words. We adopted the list provided by AI Singapore, which includes both English and Thai terms. Furthermore, we introduced a new attribute, \texttt{corrupt\_unicode}, which identifies the spans of corrupted Unicode characters in the text. These corrupted spans are frequently encountered in our dataset. In the quality filtering pipeline, these spans are replaced with empty strings. We included \texttt{corrupt\_unicode} in the C4 rules taggers for convenience.

Note that while the primary attribute of this rule is \texttt{lines\_with\_no\_ending\_punctuation}, we did not utilize it in our pipeline due to the differing nature of sentence-ending punctuation between Thai and English languages. Specifically, Thai sentences often do not end with punctuation marks, making this attribute less applicable.

\end{document}